\documentclass[conference]{IEEEtran}
\usepackage{times}

\usepackage[numbers]{natbib}
\usepackage{multicol}
\usepackage[bookmarks=true]{hyperref}

\usepackage{graphicx}
\usepackage{caption}
\usepackage{float}
\usepackage{amsmath}
\usepackage{amsfonts}
\usepackage{amssymb}
\usepackage{colortbl}  
\usepackage{xcolor}  
\usepackage{algorithmic}
\usepackage{algorithm}
\usepackage{makecell}  
\usepackage{multirow}  
\usepackage{pifont}
\newcommand{\cmark}{\ding{51}}%
\newcommand{\xmark}{\ding{55}}%

\let\oldtwocolumn\twocolumn
\renewcommand\twocolumn[1][]{
    \oldtwocolumn[{#1}{
	\begin{center}
           \includegraphics[width=1.0\textwidth]{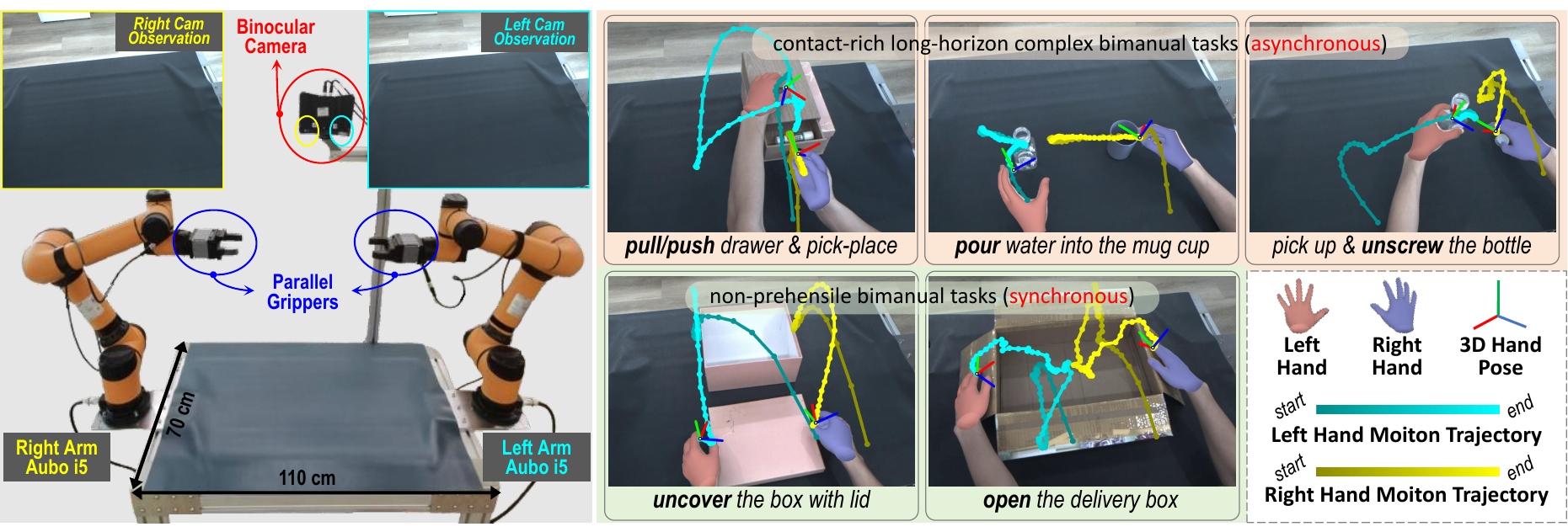}
	\captionof{figure}{Our proposed \textbf{YOTO} (\textbf{Y}ou \textbf{O}nly \textbf{T}each \textbf{O}nce) facilitates various complex long-horizon bimanual tasks. It needs only a one-shot observation of a single third-person binocular camera to extract the fine-grained motion trajectory of human hands, which can then be utilized for the dual-arm coordinated action injection and rapid proliferation of training demonstrations.}
           \label{teaser}
	\end{center}
    }]
}

\begin{document}

\title{You Only Teach Once: Learn One-Shot Bimanual Robotic Manipulation from Video Demonstrations}




%

\author{
\authorblockN{
Huayi Zhou\authorrefmark{1},
Ruixiang Wang\authorrefmark{2},
Yunxin Tai\authorrefmark{3}, 
Yueci Deng\authorrefmark{3},
Guiliang Liu\authorrefmark{1} and
Kui Jia\authorrefmark{1}\authorrefmark{4}
}
\authorblockA{
\authorrefmark{1}The Chinese University of Hong Kong, Shenzhen.\quad
\authorrefmark{2}Harbin Institute of Technology, Weihai.\quad\\
\authorrefmark{3}DexForce, Shenzhen.\quad
\authorrefmark{4}The Corresponding Author.}
}

\maketitle
\begin{abstract}

Bimanual robotic manipulation is a long-standing challenge of embodied intelligence due to its characteristics of dual-arm spatial-temporal coordination and high-dimensional action spaces. Previous studies rely on pre-defined action taxonomies or direct teleoperation to alleviate or circumvent these issues, often making them lack simplicity, versatility and scalability. Differently, we believe that the most effective and efficient way for teaching bimanual manipulation is learning from human demonstrated videos, where rich features such as spatial-temporal positions, dynamic postures, interaction states and dexterous transitions are available almost for free. In this work, we propose the YOTO (You Only Teach Once), which can extract and then inject patterns of bimanual actions from as few as a single binocular observation of hand movements, and teach dual robot arms various complex tasks. Furthermore, based on keyframes-based motion trajectories, we devise a subtle solution for rapidly generating training demonstrations with diverse variations of manipulated objects and their locations. These data can then be used to learn a customized bimanual diffusion policy (BiDP) across diverse scenes. In experiments, YOTO achieves impressive performance in mimicking 5 intricate long-horizon bimanual tasks, possesses strong generalization under different visual and spatial conditions, and outperforms existing visuomotor imitation learning methods in accuracy and efficiency. Our project link is \url{https://hnuzhy.github.io/projects/YOTO}.
\end{abstract}

\IEEEpeerreviewmaketitle

\section{Introduction}

Bimanual manipulation is an enduring topic in the robotics community \cite{balakrishnan1999exploring, peer2007towards, smith2012dual, hebert2013dual, xie2020deep, chitnis2020efficient, drolet2024comparison}. It has been widely involved in many other fields such as bionics, high-end manufacturing, mechanical control, reinforcement learning and computer vision. Despite this, achieving efficient, precise and robust manipulation of dual-arm robots to accomplish various daily tasks remains a difficult research area. Generally, there are two main challenges: coordination and state complexity \cite{grotz2024peract2, liu2024voxact}. On the one hand, the two arms working together need to move alternately or simultaneously in a coordinated, non-procrastinated manner and avoid collisions with the scene or each other. This places stringent demands on the control and scheduling scheme. On the other hand, the total degrees of freedom of two arms and their respective end effectors are distributed in a higher-dimensional space than a single arm. This makes the design of motion planning and action prediction more challenging. Given these difficulties, it is no small feat to drive two robot arms to perform tasks that human toddlers can do with ease, such as uncovering lids, assembling blocks and lifting large-size objects, let alone mastering many more complex long-horizon skills.

\begin{figure*}
	\begin{center}
           \includegraphics[width=1.0\linewidth]{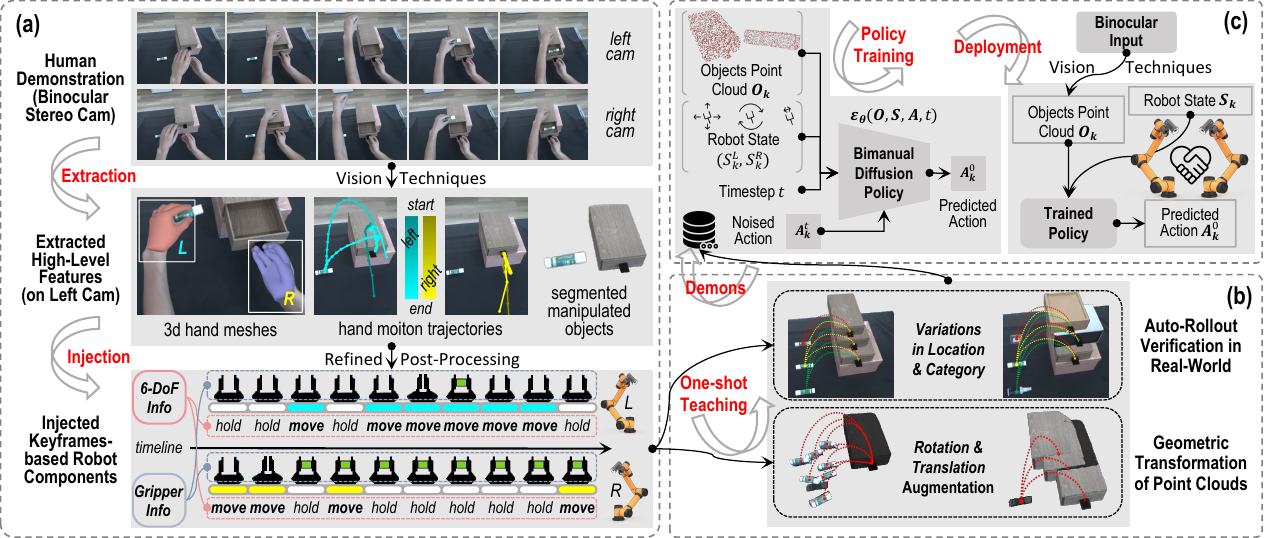}
	\vspace{-10pt}
	\captionof{figure}{The overview of our proposed YOTO. It is a general framework consists of three main modules: \textbf{(a)} the human hand motion extraction and injection, \textbf{(b)} the training demonstration proliferation from one-shot teaching, and \textbf{(c)} the training and deployment of a customized bimanual diffusion policy (BiDP). It is best to zoom in to view the details.}
           \label{framework}
	\vspace{-15pt}
	\end{center}
\end{figure*}

The mainstream bimanual manipulation research includes two major branches: explicitly classifying tasks based on pre-defined taxonomy \cite{smith2012dual, krebs2022bimanual, grotz2024peract2, liu2024voxact} and implicitly learning from demonstrations collected by teleoperation \cite{zhao2023learning, o2024open, mu2024robotwin, liu2024rdt}. The former often fails to uniformly cover arbitrary tasks and also limits the flexibility of the robot arm. While the latter requires substantial training data which is inconvenient to scale up. And collected demonstrations are intrinsically non-stationary and despatialized, which is not conducive to training robust and generalizable action policies. In addition to taxonomy and teleoperation, an indirect but more plausible and interpretable route is to learn from human action videos \cite{bahety2024screwmimic, zhou2024learning, gao2024bi,  chen2024object, li2024okami, peng2024tiebot, kerr2024robot}. This route is based on relatively mature vision techniques to process human demonstrations and extract high-level features for generating robot manipulation-relevant elements. In this paper, we also follow this promising path. Our dual-arm workbench, hardware settings, and selected bimanual tasks are shown in Fig.~\ref{teaser}. And the overall framework is shown in Fig.~\ref{framework}. 

Specifically, we focus on understanding human hands, including their location, left-rightness, 3D shape, joints, pose, contact, and open/closed state. These features can be perceived using hand-related vision methods \cite{potamias2024wilor, shan2020understanding, pavlakos2024reconstructing}. After extracting hand motion trajectories, we do not simply inject step-wise actions into robots. Because visual perception results are inevitably erroneous, and real hand motions are jittery and discontinuous. We thus simplify the consecutive trajectory into discrete keyframes \cite{james2022coarse, shridhar2023perceiver}, and assign the corresponding keyposes to two arms to execute by applying inverse kinematics interpolation. Besides, we also record and replay the order of dual-hand movements (termed as \textit{motion mask}), which can help to address the dual-arm coordination issue in long-horizon bimanual tasks. Now, we successfully obtain a stable and refined manipulation motion exemplar. 

More than that, thanks to the editability of obtained single teaching, we devise rapid proliferation strategies of training demonstrations. First, we change the 6-DoF pose of task-related objects and adjust corresponding keyposes to let real robots replay similar actions. Objects can also be replaced with other ones of analogous shape and size. This auto-rollout operation is stable and much faster than teleoperation \cite{zhao2023learning, wang2024dexcap}. For example, we can collect about 300 demonstrations in 8 hours based on a well-taught task. On the other hand, after knowing the reachable area of manipulators, we can perform geometric transformation on segmented object point clouds, which can be extracted by using open vocabulary segmentation \cite{xiao2024florence, ravi2024sam} and binocular stereo matching \cite{xu2023iterative, xu2024igev}. Such augmentation is more reliable and efficient than rollout. Therefore, mixing the above two data expansion schemes, we call it proliferation, just like the generation of cells.

With sufficient training data, we follow diffusion-based visuomotor imitation methods \cite{chi2023diffusion, ze2024dp3, yange2024quibot} and propose a specialized bimanual diffusion policy (BiDP), which is customized for learning long-horizon dual-arm tasks. It has three major improvements. First, we reduce observations (\textit{e.g.}, 3D point clouds) from the entire scene to manipulated objects to accelerate training convergence and eliminate irrelevant terms \cite{goyal2024rvt, liu2024voxact}. Then, instead of modeling continuous actions, we choose to predict essential keyposes \cite{ma2024hierarchical, xian2023chaineddiffuser, ke20243d, zeng2024learning}, which can greatly decrease the diffusion space dimensionality. Third, we utilize the motion mask to determine the alternating or synchronous dual-arm moving, and reorganize the bimanual action space to train a unified action policy. In experiments, we have verified the high efficiency and effectiveness of BiDP on challenging bimanual tasks. 

Overall, we have the following four contributions:

\begin{itemize}
\item We present a paradigm for extracting and injecting dual-arm movements from a one-shot observation of human hands demonstration, which supports the fast transfer of bimanual manipulation skills to two robotic arms.
\item We develop a solution for rapidly proliferating training demonstrations based on one-shot teaching, which is more convenient and reliable than teleoperation.
\item We propose a dedicated bimanual diffusion policy (BiDP) algorithm that can efficiently and effectively assist dual-arm manipulators in imitating complex skills.
\item Our framework YOTO is compatible with most bimanual tasks. We verified its effectiveness and superiority on 5 complex long-horizon manipulation tasks (including synchronous and asynchronous).
\end{itemize}

\section{Related Works}

\subsection{Bimanual Robotic Manipulation}
Many bimanual manipulation methods focus on specialized tasks or primitive skills, such as cloth-folding \cite{maitin2010cloth, bersch2011bimanual, colome2018dimensionality, weng2022fabricflownet, avigal2022speedfolding, canberk2023cloth, salhotra2023learning}, bagging \cite{chen2023autobag, chen2023bagging, bahety2023bag}, untangling \cite{grannen2021untangling, peng2024tiebot}, untwisting \cite{lin2024twisting, bahety2024screwmimic}, throwing/catching \cite{huang2023dynamic, yan2024impact, li2023efficient}, scooping \cite{grannen2023learning}, carrying \cite{sirintuna2023carrying} and dressing \cite{zhu2024you}. For general bimanual manipulation, typical research \cite{smith2012dual, hebert2013dual, mirrazavi2016coordinated, krebs2022bimanual, hartmann2022long, zhaodual2023afford, yang2024asymdex} tends to explicitly classify them into uncoordinated and coordinated, or symmetrical and asymmetrical according to task characteristics. Some homologous approaches assume that two arms form a leader-follower \cite{liu2022robot, grotz2024peract2} or stabilizer-actor \cite{grannen2023stabilize, liu2024voxact} pair. Most recently, the ALOHA series \cite{zhao2023learning, fu2024mobile, aldaco2024aloha, zhao2024aloha} have revolutionized bimanual manipulation by dexterous teleoperating and upgrading low-cost hardwares of real-world robotics. These similar works \cite{zhao2023learning, team2024octo, kim2024openvla, mu2024robotwin, liu2024rdt, black2024pi0} implicitly train an end-to-end imitation network using massive and diverse teleoperated data, expecting to get generalized large robotic models. To further improve dual-arm reachability and dexterity, some studies have equipped multi-finger hands \cite{lin2024learning, wang2024dexcap, shaw2024bimanual, ding2024bunny, cheng2024open, fu2024humanplus}, mobile footplates \cite{yang2024equivact, yange2024quibot, zhang2024empowering, fu2024mobile}, tactile feedbacks \cite{lin2024learning, ding2024bunny, chen2024arcap} or active cameras \cite{chuang2024active, cheng2024open}.  In contrast to them, our manipulators are two fixed-base robot arms with parallel-jaw grippers. We propose an universal framework that learns bimanual policies with considering the dual-arm coordination. And the training data is not collected via teleoperation but proliferated from a single-shot demonstration. 

\subsection{Learn from Human Hand Videos}
Human hand videos are valuable resources for learning complex manipulation behaviors \cite{grauman2022ego4d, fan2023arctic, zhan2024oakink2, liu2024taco, grauman2024ego}. Extensive research has leveraged human demonstrations to learn robot manipulation by extracting rich non-privileged features, such as keypoints \cite{papagiannis2024rx, gao2024bi, wen2023any}, affordances \cite{li2024learning, zhang2024affordance, nasiriany2024rt}, 3D hand poses \cite{li2024okami, kerr2024robot, bahety2024screwmimic}, motion trajectories \cite{li2024okami, kerr2024robot, chen2024object, zhang2024affordance} and invariant correspondences \cite{peng2024tiebot, ko2024learning, zhang2024one}. These features can be tailored to robot-specific variables to alleviate morphology gaps, such as manipulation plans, retargeted motions and precise actions. Two contemporary works \cite{kerr2024robot, li2024okami} also propose to use a single human demonstration to learn bimanual manipulation similar to us. RSRD \cite{kerr2024robot} roughly recovers 3D part motion of articulated objects from a monocular RGB video, while we adopt a binocular camera to more accurately capture arbitrary object in 3D space. OKAMI \cite{li2024okami} applies the object-aware motion retargeting which is noisy and non-smooth, while we devise a keyframes-based motion extraction scheme which is more robust and versatile.


\subsection{Visuomotor Imitation Learning}
Visuomotor imitation learning aims to train action prediction policies based on visual observations by exploiting labeled demonstrations \cite{mandlekar2020learning, johns2021coarse, mandlekar2022matters, jang2022bc, wang2023mimicplay, james2022coarse, shridhar2023perceiver}. These learned policies can drive robots to complete various manipulation with just dozens of demonstrations, covering long-horizon \cite{mandlekar2020learning, wang2023mimicplay}, dexterous \cite{ze2024dp3, wang2024dexcap} and bimanual \cite{zhao2023learning, shaw2024bimanual} tasks. Especially, \citet{zhao2023learning} introduced the action chunking transformers (ACT) to learn high-frequency controls with closed-loop feedback in an end-to-end manner. \citet{chi2023diffusion} adopted conditional denoising diffusion models \cite{ho2020denoising, song2021denoising, nichol2021improved} to represent visuomotor policies in robotics, exhibiting impressive training stability in modeling high-dimensional action distributions. \citet{ze2024dp3} incorporated 3D conditioning into the original diffusion policy \cite{chi2023diffusion}, rather than focusing on RGB images and states as conditions. \citet{yange2024quibot} combined SIM(3)-equivariance \cite{yang2024equivact, ryu2024diffusion, brehmer2024edgi} with diffusion policy, acquiring a more generalizable and sample-efficient visuomotor policy than \cite{chi2023diffusion, ze2024dp3}. Inspired by them, we propose a bimanual diffusion policy (BiDP), which adds motion mask as a new diffusion condition and simplifies visual observations to task-related object point clouds, making it suitable for learning bimanual manipulation tasks.

\section{Hardware System}

\textbf{Dual-Arm Placement:} 
Most human video-inspired bimanual manipulation works apply humanoid robots \cite{bahety2024screwmimic, gao2024bi, li2024okami, peng2024tiebot, kerr2024robot} or two ipsilateral arms \cite{zhou2024learning, gao2024bi} to build workstations. Some bimanual teleoperations also tend to be anthropopathic \cite{lin2024learning, shaw2024bimanual, cheng2024open, fu2024humanplus} or ipsilateral \cite{wang2024dexcap, chen2024arcap, ding2024bunny}. Despite the similarity to human morphology, they are not necessarily optimal. Comparatively, it is possible to place two manipulators opposite each other, as in ALOHA series \cite{zhao2023learning, fu2024mobile, aldaco2024aloha, zhao2024aloha} and its followers \cite{chuang2024active, liu2024rdt, black2024pi0}. This heterolateral setup minimizes the overlap of accessible space and is thus compatible with a wider range of bimanual tasks. We also adopt the contralateral placement as shown in the left of Fig.~\ref{teaser}, where each arm (Aubo i5\footnote{https://www.aubo-cobot.com/public/i5product3}) has a span of approximately 880 mm.

\textbf{End Effector Selection:} 
Although some methods utilize multi-fingered dexterous hands as end effectors \cite{wang2024dexcap, shaw2024bimanual, cheng2024open, fu2024humanplus} and even add tactile sensors \cite{lin2024learning, ding2024bunny, chen2024arcap} to the hands, we still use two parallel-jaw grippers (with max opening distance 80 mm of each DH-Robotics\footnote{https://en.dh-robotics.com/product/pgi}), which are easier to control and interpret. We will show that it is sufficient to complete complex tasks that are non-prehensile or synchronous.

\textbf{Camera Observation:}  
Many previous methods adopt the multi-view RGB observations \cite{zhao2023learning, liu2024rdt, black2024pi0}, mainly including the global third-person camera and the local eye-in-hand camera. Other works have shown that a single third-person RGB-D camera \cite{wang2024dexcap, bahety2024screwmimic, gao2024bi, li2024okami} is also acceptable. We use a binocular stereo camera (the DexSense 3D industrial camera\footnote{https://dexforce-3dvision.com/productinfo/1022811.html}), similar to commercial RGB-D cameras, but providing raw left and right images to enable flexible post-processing.

\section{Method}

In this part, we introduce in detail the proposed framework YOTO, which contains three major modules and is illustrated in Fig.~\ref{framework}. We firstly give a basic definition of the problem in Sec.~\ref{mPF}. Then, a detailed explanation of the three core modules is presented, which includes the standardized hand motion extraction and injection process in Sec.~\ref{mHMEI}, the demonstration proliferation solution from one teaching in Sec.~\ref{mDPOT} and the proposed visuomotor bimanual diffusion policy (BiDP) method in Sec.~\ref{mBDPL}.

\subsection{Problem Formulation}\label{mPF}
In this paper, we mainly consider bimanual robot manipulation tasks, where the agent (e.g., dual manipulators equipped with parallel-jaw grippers) does not have access to the ground-truth state of the environment, but visual observations $O$ from a binocular camera and robots proprioception states $S$. As for the action space $A\!=\!\{a^{p}\in\mathbb{R}^3, a^{r}\in\mathbb{SO}(3), a^{g}\in\{0,1\}\}$, it includes the target 6-DoF pose of each robot arm and the binary open/closed state of the gripper. Note, we focus on bimanual tasks sharing the same observations $O$. For the chirality, we utilize $\diamond\!\in\!\{L,R\}$ to distinguish two robot arms, such as $S^L$, $S^R$, $A^L$ and $A^R$. The same applies to the difference between left and right hands below.

For imitation learning, the agent mimics manipulation plans from labeled demonstrations $\mathcal{D}\!=\!\{(\mathbf{O}, \mathbf{A})_i\}_{i=1}^N$, where $N$ is the number of trajectories, $\mathbf{O}\!=\!\{O_t, S^L_t, S^R_t\}_{t=1}^T$ are observations of all $T$ steps, and $\mathbf{A}\!=\!\{A^L_t, A^R_t\}_{t=1}^T$ are actions to complete the task. The learning objective can be simply concluded as a maximum likelihood observation-conditioned imitation objective to learn the policy $\pi_\theta$:
\begin{equation}
	\ell = \mathbb{E}_{(\mathbf{O}, \mathbf{A})_i \sim \mathcal{D}}\left[\sum\nolimits^{|\mathbf{O}|}_{t=0}\log{\pi_\theta}(A^{\diamond}_t | O_t, S^{\diamond}_t) \right].
	\label{eqnA}
\end{equation}

Next, we present how to obtain sufficient training demonstrations proliferated from only a single-shot human teaching and how to improve existing diffusion-based imitation policies for addressing the bimanual manipulation problem.

\subsection{Hand Motion Extraction and Injection}\label{mHMEI}
This part corresponds to the module in Fig.~\ref{framework} \textbf{(a)}.We first manually demonstrate a long-horizon bimanual task using two hands on the dual-arm accessible operating table. Then, we leverage favourable vision techniques to extract rich manipulation features from recorded videos by a single binocular camera. Extracted features will be post-processed to obtain keyframes-based motion variables (such as 6-DoF poses and gripper states) that can drive dual arms.


\subsubsection{Human Demonstration Capturing}
By default, we capture dual-stream synchronized RGB videos with slight necessary visual difference between left and right cameras to estimate disparity and depth map. We mainly observe the left RGB view to extract a series of hand-related features, and thus always keep both hands visible to the left camera. The right view is only awakened when accurate 3D information is needed in a particular frame. This reduces the computational burden of stereo matching \cite{xu2023iterative} by at least half.

\subsubsection{High-Level Features Extraction}
Given a video demonstration (the left stream) of one specified bimanual task, we run our vision perception pipeline to obtain the 3D point trajectories and status of two hands.

\textbf{3D point trajectories.} We first use WiLoR \cite{potamias2024wilor} to detect bounding boxes of left and right hands in each frame and then estimate their 3D shapes $\mathcal{H}^L$ and $\mathcal{H}^R$ represented by MANO \cite{romero2017embodied}. Then, we simply track the center point $h^{p,\diamond}_j\!=\!(x^{\diamond}_j, y^{\diamond}_j, z^{\diamond}_j)$ of each hand and obtain the 3D hands sequence $\mathbf{H}\!=\!\{(\mathcal{H}^{\diamond}_j, h^{p,\diamond}_j)\}^{J}_{j=1}$, where $\diamond$ is the chirality and $j$ is the index among all $J$ frames. The $h^{p,\diamond}_j$ can be calculated by averaging several selected points (\textit{e.g.,} five finger tips) from 21 pre-defined joints of the 3D hand model $\mathcal{H}^{\diamond}_j$.

As of here, many similar works \cite{kerr2024robot, li2024okami, cheng2024open, fu2024humanplus} choose to retarget the produced continuous trajectories $\{h^{p,\diamond}_j\}^{J}_{j=1}$ to their end effectors through estimated 3D geometric transformations. However, considering the inherent errors of hand-related vision algorithms in left-right classification and 3D shape regression, we cannot fully trust trajectories directly derived from them. In particular, current state-of-the-art 3D hand mesh reconstruction methods, such as WiLoR \cite{potamias2024wilor} and HaMeR \cite{pavlakos2024reconstructing}, still cannot achieve continuous and consistent prediction in a given camera space. This is also pointed out and verified by DexCap \cite{wang2024dexcap}. More examples can be found in Fig.~\ref{motions}. As an alternative, we propose to project all 3D points $\{h^{p,\diamond}_j\}^{J}_{j=1}$ onto the 2D image, and then lift these points to 3D by applying the stereo matching algorithm \cite{xu2023iterative}. The final back-projected 3D point trajectories are $\{\widehat{h}^{p,\diamond}_j\}^{J}_{j=1}$, which are guaranteed to be more stable in the given camera space.

\begin{algorithm}[t]
\caption{3D Hand Pose Calculation.}  
\begin{algorithmic}\small 
	\STATE $\bullet$ \textbf{Input:} 3D hand shapes $\mathcal{H}^{\diamond}_j$, index array of 21 pre-defined 3D hand joints $I_\text{hand}$, index numbers of wrist joint $i_\text{wri}$ / index-fingertip $i_\text{ind}$ / ring-fingertip $i_\text{ring}$, the given chirality $\diamond = L$ or $\diamond = R$.
	\STATE $\bullet$ \textbf{Output:} 3D hand poses $h^{r,\diamond}_j$.  \hfill\textcolor{olive}{// either $L$ or $R$}
	\STATE Initialize $\mathbf{P}^{\diamond}_j \leftarrow \textbf{MANO}(\mathcal{H}^{\diamond}_j, I_\text{hand})$; \hfill\textcolor{olive}{// 3D hand joints indexing}
	\STATE $p_\text{wri} \leftarrow \mathbf{P}^{\diamond}_j[i_\text{wri}], \; p_\text{ind} \leftarrow \mathbf{P}^{\diamond}_j[i_\text{ind}], \; p_\text{ring} \leftarrow \mathbf{P}^{\diamond}_j[i_\text{ring}]$;
	\STATE $l_\text{iw} \leftarrow (p_\text{ind} - p_\text{wri}), \; l_\text{rw} \leftarrow (p_\text{ring} - p_\text{wri})$; \hfill\textcolor{olive}{// two 3D lines}
	\STATE $v_z \leftarrow \textsc{cross\_product}(l_\text{iw}, l_\text{rw})$; \hfill\textcolor{olive}{// Z-axis direction}
	\STATE $\bar{v}_z \leftarrow v_z / (\textsc{normalize}(v_z) + \text{1e-8})$; \hfill\textcolor{olive}{// vector normalization}
	\STATE $v_y = l_\text{mid} \leftarrow (l_\text{iw} + l_\text{rw})/2.0$; \hfill\textcolor{olive}{// middle line (Y-axis direction)}
	\STATE $\bar{v}_y \leftarrow v_y / (\textsc{normalize}(v_y) + \text{1e-8})$; \hfill\textcolor{olive}{// vector normalization}
	\STATE $\bar{v}_x \leftarrow \textsc{cross\_product}(\bar{v}_y, \bar{v}_z)$; \hfill\textcolor{olive}{// X-axis direction}
	\STATE $v_{rot} \leftarrow \textsc{concatenate}([\bar{v}_x, \bar{v}_y, \bar{v}_z])$; \hfill\textcolor{olive}{// final $3\!\times\!3$ rotation matrix}
	\STATE \textbf{return} $v_{rot}$;
\end{algorithmic}
\label{algHandPose}
\end{algorithm}

\textbf{States of two hands.} In order to fully map hand movements to two-fingered grippers, we also need to determine the 3D orientations $h^{r,\diamond}_j$ and open/closed states $h^{g,\diamond}_j$ by further observing 3D hands $\mathcal{H}^{\diamond}_j$. Here, we can estimate the open/closed state by detecting if the hand is in contact with an object \cite{shan2020understanding}. If there is contact, the hand is considered closed ($h^{g,\diamond}_j\!=\!0$), otherwise open ($h^{g,\diamond}_j\!=\!1$). This is more trustworthy than relying solely on hands to estimate status. For calculating 3D hand poses $h^{r,\diamond}_j$, we need to simplify the hand into a lower-dimensional gripper, which is analogous to the eigengrasping \cite{miller2004graspit, ciocarlie2007dexterous}. We summarize this process in Alg.~\ref{algHandPose}. To this point, we have obtained the rough motion trajectories purely based on human hand videos $\{(\widehat{h}^{p,\diamond}_j, h^{r,\diamond}_j, h^{g,\diamond}_j)\}^{J}_{j=1}$.

Additionally, we adopt cutting-edge vision algorithms (including the vision-language model Florence-2 \cite{xiao2024florence} and SAM2 \cite{ravi2024sam}) to extract segmented manipulated objects from the left initial image as our disturbance-free visual observations $\widehat{O}$, which will be further lifted to 3D point clouds $\widetilde{O}$ by applying stereo matching approaches \cite{xu2023iterative, xu2024igev}.

\subsubsection{Robot Actions Injection}\label{mHMEI3}
Although we have obtained robot-oriented motion trajectories, their validity and usability are still concerns. For example, some target poses may be unreachable for the failed inverse kinematics. Due to agnostic structures, two arms may collide at some point. An obvious approach is to replay and verify the rationality of each action step by step directly on real robots, but this choice is unsafe and inefficient, considering that the total number of frames $J$ is usually about 100 to 200. 

\begin{figure}
	\begin{center}
           \includegraphics[width=1.0\linewidth]{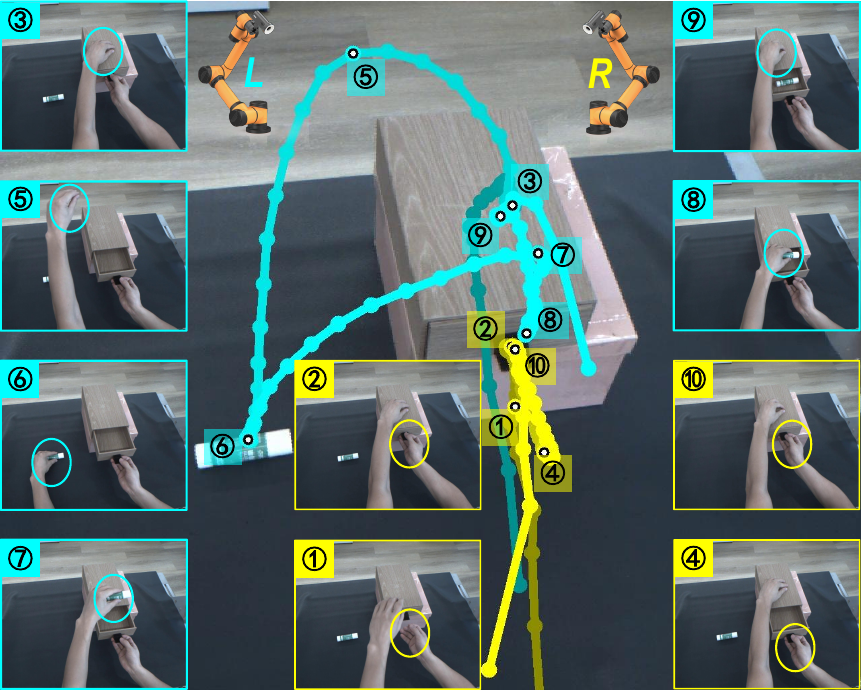}
	\vspace{-10pt}
	\captionof{figure}{A detailed example of extracted motion trajectories with corresponding keyframes of both left hand and right hand. It is best to zoom in to view the details.}
           \label{keyframes}
	\vspace{-15pt}
	\end{center}
\end{figure}

\textbf{Keyframes-based motion actions.} To this end, we turn to a more reasonable and safer post-processing, namely keyframes-based motion simplification and injection. Specifically, we inherit the abstraction of a consequent demonstration into discrete keyframes (\textit{a.k.a.} keyposes) as in C2FARM \cite{james2022coarse} and PerAct \cite{shridhar2023perceiver}. Keyframes are important intermediate end-effector poses that summarize a demonstration and can be auto-extracted using simple heuristics, such as a change in the open/close end-effector state or local extrema of velocity/acceleration. This concept is widely used in long-horizon manipulation studies \cite{ma2024hierarchical, xian2023chaineddiffuser, ke20243d, zeng2024learning}. Accordingly, we can just learn to predict the next best keyframe, and use a sampling-based motion planner to reach it during inference. We thus simplify trajectories $\{(\widehat{h}^{p,\diamond}_j, h^{r,\diamond}_j, h^{g,\diamond}_j)\}^{J}_{j=1}$ into a set of keyframes $\{(\widetilde{h}^{p,\diamond}_k, \widetilde{h}^{r,\diamond}_k, \widetilde{h}^{g,\diamond}_k)\}^{K}_{k=1}$, where $k$ is the index of $K$ keyframes. $K$ is around 10 in our tasks ($K\!\ll\!J$), which makes it much more easier to quickly verify and correct errors. To inject these keyposes into the dual-arm robot, we need to transform them from the camera coordinate to the robot coordinate using the pre-measured hand-eye calibration transformation matrix. Usually, a real-robot verification takes about three minutes. We finally update the verified trajectories into $\widetilde{\mathbf{A}}\!=\!\{(\widetilde{a}^{p,\diamond}_k, \widetilde{a}^{r,\diamond}_k, \widetilde{a}^{g,\diamond}_k)\}^{K}_{k=1}$, which consists of the successfully injected $K$ robot actions. An elaborate example of extracted keyframes is shown in Fig.~\ref{keyframes}.

\textbf{Derivation of motion mask.} Additionally, we should always care about the dual-arm spatial-temporal coordination, which is one of the core issues of bimanual manipulation. Fortunately, when we extract the hand motions, we already have a time record in every frame, which represents the refined keyframes-based set $\widetilde{\mathbf{A}}$ naturally contains detailed timestamps. Based on it, we can thus derive the corresponding coordination strategy $\mathbf{C}\!=\!\{(\mathcal{C}^L_k, \mathcal{C}^R_k) | \mathcal{C}^{\diamond}_k\!\in\!\{0,1\} \}^{K}_{k=1}$, where $\mathcal{C}^{\diamond}_k$ means the motion state of a robot arm at the $k$-th keyframe. The binary value 0 means holding on, 1 means moving on. Given this particularity, we name it \textit{motion mask} to schedule robot motion. A specific illustration of $\mathbf{C}$ for the pull drawer task can be found in the down-left corner of Fig.~\ref{framework}. This example is broadly applicable to strictly asynchronous bimanual tasks (\textit{e.g.,} $\mathcal{C}^L_k \neq \mathcal{C}^R_k$). While, for fully synchronous manipulation tasks, values of $\mathcal{C}^{L}_k$ and $\mathcal{C}^{R}_k$ in $\mathbf{C}$ keep the same. Currently, we do not consider those long-horizon tasks where synchronized and asynchronized keyframes are mixed.

In the following, we show that the extracted fine-grained keyframes-based motion actions $\widetilde{\mathbf{A}}$ along with the corresponding \textit{motion mask} $\mathbf{C}$ will continue to play a vital role.

\subsection{Demonstration Proliferation from One Teaching}\label{mDPOT}
Based on the one-shot teaching, we propose two demonstration proliferation schemes, the automatic rollout verification of real robots and point cloud-level geometry augmentation of manipulated objects. This solution is an efficient and reliable route to quickly produce training data for imitation learning. An example is shown in Fig.~\ref{framework} \textbf{(b)}.

\subsubsection{Auto-Rollout Verification in Real-World}
Formally, our refined keyframes-based robot actions $\widetilde{\mathbf{A}}$ are interpretable and editable. These properties assist us to conduct automated demonstration rollout verification and collection on real robots. First, we can easily split $\widetilde{\mathbf{A}}$ into two distinctive trajectories $\widetilde{\mathbf{A}}^{L}$ and $\widetilde{\mathbf{A}}^{R}$ belonging to the left and right robotic arms based on the motion mask $\mathbf{C}$. Below is for decomposing strictly asynchronous tasks.
\begin{equation}
	\left\{ \begin{array}{rcl}
	\widetilde{\mathbf{A}}^{L} &=& \{(\widetilde{a}^{p, L}_k, \widetilde{a}^{r, L}_k, \widetilde{a}^{g, L}_k) | \mathcal{C}^{L}_k=1, \mathcal{C}^{R}_k=0\}, \\
	\widetilde{\mathbf{A}}^{R} &=& \{(\widetilde{a}^{p, R}_k, \widetilde{a}^{r, R}_k, \widetilde{a}^{g, R}_k) | \mathcal{C}^{L}_k=0, \mathcal{C}^{R}_k=1\}, \\
	K &=& |\widetilde{\mathbf{A}}^{L}| + |\widetilde{\mathbf{A}}^{R}| \;\;=\;\; |\widetilde{\mathbf{A}}| / 2, \\
	\end{array}\right.
	\label{eqnB}
\end{equation}
where we actually eliminate $K$ redundant keyposes for unilateral arm waiting (holding on actions). For synchronous tasks ($|\widetilde{\mathbf{A}}^{L}|\!=\!|\widetilde{\mathbf{A}}^{R}|\!=\!K$), we always have to drive both arms, so there is no need to apply the motion mask.

The above allows two arms to disengage smoothly. Then, we can precisely edit any keyframe in $\widetilde{\mathbf{A}}^{L}$ or $\widetilde{\mathbf{A}}^{R}$ closely related to the manipulated object to align with its changed keypose in real-world. We still take the pull drawer task (with 10 keyframes) as an example. When moving the object picked up by the left arm, we need to adjust the $6$-th keypose $\widetilde{a}^{L}_6=(\widetilde{a}^{p, L}_6, \widetilde{a}^{r, L}_6, \widetilde{a}^{g, L}_6)$. For example, if we move the object 5 cm along the X-axis positive direction, we then just add an offset $(0.05, 0.00,0.00)$ to the position part $\widetilde{a}^{p, L}_6$. Moreover, we can also replace objects with similar shapes in the same position to expand category diversity. Finally, we conduct the rollout to get a new demonstration. The same is true for adjusting the drawer manipulated by the right arm. Regardless of simplicity, we compared auto-rollout with two popular data collection methods, master-slave arm synchronization and drag-and-drop teaching, and found that it is more efficient. See Tab.~\ref{tabA} for the comparison. The other two ways are hampered by multi-operators and higher failure rates.

\subsubsection{Geometric Transformation of Point Clouds}
Regarding the above expansion of object positions and categories in real-world, we still have to verify them one by one. We thus expect to reliably augment visual observations of manipulated objects (the extracted 3D point clouds $\widetilde{O}$) any number of times, so that theoretically infinite demonstrations can be obtained. In the auto-rollout stage, we have initially figured out the correspondence between manipulated objects and their relevant keyframes. Now, we can perform geometric transformations (mainly controlled rotations and translations) on the objects at the point cloud level, and update the 6-DoF values in the corresponding keyframes. In this way, matching pairs of visual observations $\widetilde{O}$ and keyframes-based actions $\widetilde{\mathbf{A}}$ can be generated in batches, forming a series of new training data, which no longer need to be verified in real robots. It should be noted that the geometric transformation of $\widetilde{O}$ is restricted, that is, it cannot exceed the reach of the robot arm. Fortunately, the rational moving range of manipulated objects can be measured during the auto-rollout phase incidentally. In Tab.~\ref{tabA}, we have added the time comparison of this data proliferation, which maintains the highest efficiency.

\setlength{\tabcolsep}{4pt}
\begin{table}[!t]  
	\centering
	\caption{The time comparison of different data collection or expansion methods. We report the average completion time for 3 tasks, 10 valid trials in total for each task. The $\dagger$ means it can be achieved by directly modifying the script.}
	\begin{tabular}{l|c|c|ccc}
	\Xhline{1.2pt}
	\multirow{3}{*}{Methods} & \multirow{3}{*}{Operators} & \multirow{3}{*}{Arms} & \multicolumn{3}{c}{Long-Horizon Bimanual Tasks} \\
	\cline{4-6}
	~ & ~ & ~ & \multirow{2}{*}{\makecell{pull\\drawer (s)}} & \multirow{2}{*}{\makecell{pour\\water (s)}} & \multirow{2}{*}{\makecell{unscrew\\bottle (s)}} \\  
	~ & ~ & ~ & ~ & ~ & ~ \\
	\Xhline{1.2pt}
	Master-Slave & 2 & 2 & 204.8 & 226.2 & 247.9 \\
	Drag\&Drop & 2 & 1 & 100.7 & 115.4 & 123.6 \\
	Auto-Rollout & 1 & 1 & 41.5 & 52.1 & 51.4 \\
	Geo-Trans $\dagger$ & 1 & 0 & 1.5 & 1.5 & 1.0 \\
	\Xhline{1.2pt}
	\end{tabular}
	\label{tabA}
	\vspace{-10pt}
\end{table}

\subsection{Bimanual Diffusion Policy Learning}\label{mBDPL}
In this part, we adapt popular visuomotor diffusion policies \cite{chi2023diffusion, ze2024dp3, yange2024quibot}, and propose a customized bimanual diffusion policy (BiDP) to enable fast and robust imitation of long-horizon tasks. We firstly shrink the input observations into task-relevant object point clouds, allowing the policy model to converge quickly and resistant to interference. Additionally, we devise a motion mask to unify the action prediction and address the dual-arm coordination problem.

\textbf{Bimanual dataset composition.}
According to the definition in Sec.~\ref{mPF}, we rewrite the training set as $\widetilde{\mathcal{D}}\!=\!\{(\widetilde{\mathbf{O}}, \widetilde{\mathbf{A}}, \mathbf{C})_i\}_{i=1}^N$, where $N$ is the number of demonstrations. $\mathbf{C}$ is the motion mask containing coordination strategies. $\widetilde{\mathcal{D}}$ is generated by applying our proposed data proliferation solution to expand the seeding one-shot teaching to get a large dataset with hundreds or thousands of trajectories. Here, we update $\widetilde{\mathbf{O}}\!=\!\{\widetilde{O}_k, S^L_k, S^R_k\}_{k=1}^K$ and $\widetilde{\mathbf{A}}\!=\!\{(\widetilde{a}^{p,\diamond}_k, \widetilde{a}^{r,\diamond}_k, \widetilde{a}^{g,\diamond}_k)\}_{k=1}^K$, where $\widetilde{O}_k$ is the observation containing 3D point clouds of manipulated objects instead of the entire RGB image \cite{chi2023diffusion} or point clouds scene \cite{ze2024dp3, yange2024quibot}. $S^L_k$ and $S^R_k$ are robot proprioception states with similar formats as actions $S^{\diamond}_k\!=\!(\widetilde{s}^{p,\diamond}_k, \widetilde{s}^{r,\diamond}_k, \widetilde{s}^{g,\diamond}_k)$. $\widetilde{\mathbf{A}}$ have discrete keyposes, rather than continuous and dense robot states. Learning to predict keyposes is common in robotic manipulation \cite{ma2024hierarchical, xian2023chaineddiffuser, ke20243d, zeng2024learning}. The  policy needs to learn a mapping from the initial observation $\widetilde{O}_1$ to all subsequent keyposes $\widetilde{\mathbf{A}}$ for two arms. The history horizon and prediction horizon is 1 and $K$, respectively. In evaluation, the policy predicts all actions to be executed conditioned only on an one-shot observation $\{\widetilde{O}_1, S^L_1, S^R_1\}$ at first sight.

\begin{figure*}[t] 
	\begin{center}
           \includegraphics[width=\textwidth]{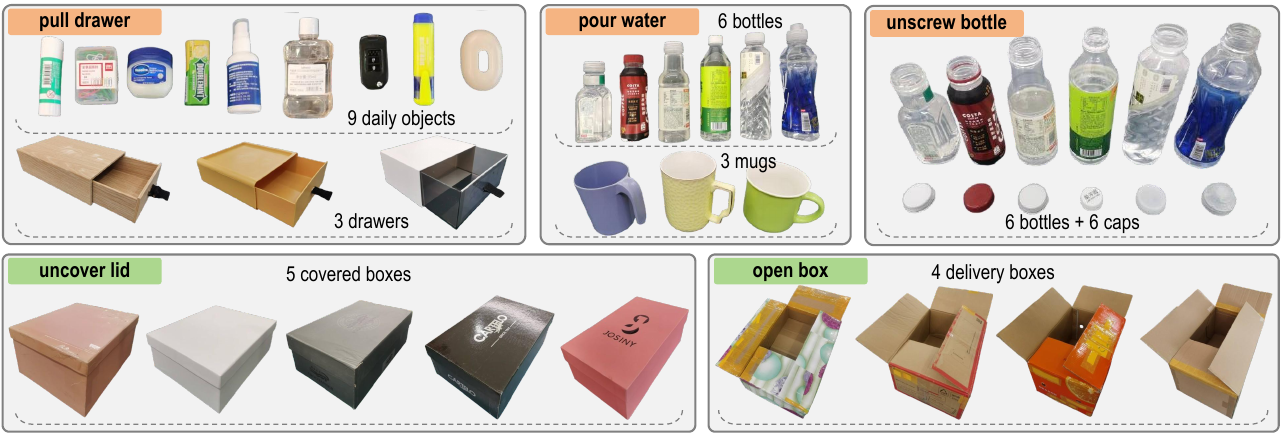}
	\vspace{-15pt}
	\captionof{figure}{We collected a variety of manipulated objects in instance-level for each of five bimanual tasks to improve and verify the generalizability of trained policies. All of these objects are from everyday life, not intentionally customized.}
           \label{assets}
	\vspace{-15pt}
	\end{center}
\end{figure*}

\textbf{Diffusion-based policy representation.}
Similar to \cite{chi2023diffusion, ze2024dp3}, we utilize Denoising Diffusion Probabilistic Models (DDPMs) \cite{ho2020denoising} to model the conditional distribution $p(\widetilde{\mathbf{A}}_k | \widetilde{\mathbf{O}}_k)$. Starting from the random Gaussian noise $\widetilde{\mathbf{A}}^T_k$, where $T$ means diffusion steps, DDPM performs $T$ iterations of denoising to predict actions with decreasing levels of noise, gradually from $\widetilde{\mathbf{A}}^{T-1}_k$ to $\widetilde{\mathbf{A}}^0_k$. This process follows:
\begin{equation}
	\widetilde{\mathbf{A}}^{t-1}_k = \alpha ( \widetilde{\mathbf{A}}^t_k - \gamma \varepsilon_{\theta}(\widetilde{\mathbf{O}}_k, \widetilde{\mathbf{A}}^t_k, t) + \mathcal{N}(0, \sigma^2, I) ).
	\label{eqnC}
\end{equation}

The policy finally outputs $\widetilde{\mathbf{A}}^0_k$. Because point clouds are used as the visual input instead of RGB images, we adopt more robust SIM(3)-equivariant architectures \cite{yang2024equivact, yange2024quibot}, rather than policies based on CNNs \cite{chi2023diffusion} or transformers \cite{ze2024dp3}. Formally, the noise prediction network $\varepsilon_{\theta}$ takes observation $\widetilde{\mathbf{O}}_k$, noisy action $\widetilde{\mathbf{A}}_k$ and diffusion timestep $t$ as input, and predicts the gradient $\triangledown\mathbf{E}(\widetilde{\mathbf{A}}_k)$ for denoising the noisy action input. It first uses a modified PointNet-based \cite{qi2017pointnet} encoder with SIM(3)-equivariance to encode visual observations. The encoded visual features and positional embeddings of $t$ are passed to FiLM layers \cite{perez2018film}. Then, the policy network applies a convolutional U-Net \cite{ronneberger2015u} to process $\widetilde{\mathbf{A}}_k$, $t$ and the conditioned observations to predict denoising gradients. Note that $\widetilde{\mathbf{O}}_k$, $\widetilde{\mathbf{A}}_k$ and $\widetilde{\mathbf{A}}^0_k$ are processed to be invariant to scale and position. Above-mentioned FiLM layers, convolutional U-net, and other connecting layers are also modulated to be $\mathbb{SO}(3)$-equivariant. Please refer to \cite{yang2024equivact, yange2024quibot} for more details.

\textbf{Customized bimanual diffusion policy.}
Since $\widetilde{\mathbf{A}}_k$ and $S^{\diamond}_k$ contain dual-arm actions in our task, it is important to preprocess them appropriately. A vanilla approach is to predict all actions in each keyframe, including $(\widetilde{a}^{p,L}_k, \widetilde{a}^{r,L}_k, \widetilde{a}^{g,L}_k)$ and $(\widetilde{a}^{p,R}_k, \widetilde{a}^{r,R}_k, \widetilde{a}^{g,R}_k)$. This not only needs to re-splice the position, rotation, and gripper data and modify the diffusion-based policy network accordingly, but also learns redundant actions for asynchronous tasks (as pointed out in Sec.~\ref{mDPOT}), which is inefficient and error-prone. To this end, we reorganize the action space into $\overline{\mathbf{A}}=\{\widetilde{\mathbf{A}}^{L}, \widetilde{\mathbf{A}}^{R}\}$ based on the motion mask $\mathbf{C}$ according to Eqn.~\ref{eqnB}. $\overline{\mathbf{A}}$ contains a series of time-ordered single-arm actions, which is a mixture of the left and right with removing potential redundancy. Taking the pull drawer task as an example, a demonstration consists of 10 keyframes $\{\widetilde{A}^{R}_1, \widetilde{A}^{R}_2, \widetilde{A}^{L}_3, \widetilde{A}^{R}_4, \widetilde{A}^{L}_5, \widetilde{A}^{L}_6, \widetilde{A}^{L}_7, \widetilde{A}^{L}_8, \widetilde{A}^{L}_9, \widetilde{A}^{R}_{10}\}$. For synchronous tasks, the left and right sides appear alternately. In this way, we unify the policy network form of bimanual tasks, which is also compatible with single-arm. More implementation details are in supplementary materials.

\section{Experiments}

We aim to answer the following research questions. $\mathsf{Q}1$: What is the quality of our extracted hand motions? $\mathsf{Q}2$: Can the various strategies introduced in the YOTO framework enable it to better learn bimanual manipulation policies? $\mathsf{Q}3$: Do trained BiDP models generalize outside of the in-distribution domain? $\mathsf{Q}4$: Is the presented YOTO framework compatible with a variety of long-horizon complex tasks?

\subsection{Experiment Setups}

\subsubsection{Tasks}
We evaluate YOTO on five real-world bimanual tasks, including \texttt{pull drawer}, \texttt{pour water}, \texttt{unscrew bottle}, \texttt{uncover lid} and \texttt{open box}. These tasks collectively encompass two types of dual-arm collaborations: strictly asynchronous and synchronous. The manipulated objects in these tasks might be rigid, articulated, deformable or non-prehensile. They also involve many primitive skills such as pull/push, pick/place, re-orient, unscrew, revolve and lift up. Some skills must require both arms to complete. More importantly, all tasks are long-horizon, indicating that they are quite complex due to containing multiple substeps. In the following, we explain each task in brief:

$\bullet$ \texttt{pull drawer:} A drawer and a daily pocketed object. It consists of 6 substeps including stable the drawer (L), pull the drawer (R), pick up the object (L), place the object into the drawer (L), stable the drawer (L), and push the drawer (R). 
 
$\bullet$ \texttt{pour water:} A capless bottle with water and an empty mug. It consists of 6 substeps including pick up the mug (R), pick up the bottle (L), bring the mug close to the bottle (R), pour water in bottle into the mug (L), put down the bottle (L), and put down the mug (R). 

$\bullet$ \texttt{unscrew bottle:} A capped bottle with water. It consists of 5 substeps including pick up the bottle (L), bring the bottle close to the right arm (L), unscrew the cap (R), put down the cap (R), and put down the bottle (L). 
 
$\bullet$ \texttt{uncover lid:} A rectangular box with a top covered lid and no handles. It consists of 3 substeps including go to the lower middle part of the lid (LR), lift up the lid (LR), and put down the lid to one side (LR). 

$\bullet$ \texttt{open box:} A delivery box with four handleable wings. It consists of 4 substeps including go close to the two vertical wings (LR), flick open two wings (LR), go close to the two horizontal wings (LR), and flick open two wings (LR). 

The statistics of these tasks are in Tab.~\ref{tabB}, where the number of keyframes is counted based on the one-shot teaching. Examples of each task are shown in Fig.~\ref{teaser} and Fig.~\ref{visresults}. 

\setlength{\tabcolsep}{4pt}
\begin{table}[!t]  
	\centering
	\caption{Detailed statistics of five bimanual tasks. The $\dagger$ means we only count these auto-rollout demonstrations.}
	\begin{tabular}{c|ccccc}
	\Xhline{1.2pt}
	Task Names & \makecell{pull\\drawer} & \makecell{pour\\water} & \makecell{unscrew\\bottle} & \makecell{uncover\\lid} & \makecell{open\\box} \\
	\Xhline{1.2pt} 
	Is Synchronous? & \xmark & \xmark & \xmark & \cmark & \cmark \\
	\# Manipulated Objects & 2 & 2 & 1 & 1 & 1 \\
	\# Substeps & 6 & 6 & 5 & 3 & 4 \\
	\# Keyframes & 10 & 11 & 12 & 12 & 16 \\
	\hline
	Avg. Duration (s) & 42 & 53 & 51 & 27 & 35 \\
	\# Categories & 9 $|$ 3 & 6 $|$ 3 & 6 & 5 & 4 \\
	\# Demonstrations $\dagger$ & 243 & 162 & 54 & 45 & 36 \\
	\Xhline{1.2pt}
	\end{tabular}
	\label{tabB}
	\vspace{-10pt}
\end{table}

\subsubsection{Demonstrations}
Imitation learning requires sufficient training data, including diverse verified task trajectories, to learn a closed-loop action prediction policy. To this end, as described in Sec.~\ref{mDPOT}, we start from a single-shot teaching of every task and collect a considerable number of demonstrations via the proposed rapid proliferation solution. Moreover, to improve and evaluate the generalization of learned policies, we have collected multiple objects within each task. All related assets are shown in Fig.~\ref{assets}.

Specifically, we first implement the auto-rollout strategy to collect real robot data. We set 3 (for tasks \texttt{pull drawer} and \texttt{pour water}) or 9 (for the other three tasks) position variations for each manipulated object, and replace all alternatives from the assets in each position. In this way, we get training data with diverse positions and categories. The demonstration number of every task is in the last row of Tab.~\ref{tabB}, where we added statistics on their average duration. We then processed these data into the form suitable for BiDP, including extracting 3D point clouds of manipulated objects and saving the corresponding multi-step end-effector keyposes. Note that we also recorded the complete binocular video observation and continuous robot actions during each auto-rollout, so that we can reproduce mainstream policy learning methods \cite{zhao2023learning, chi2023diffusion, ze2024dp3, yange2024quibot} for comparison.  Next, we applied 3D geometric transformations to each demonstration, acting only on task-relevant object point clouds. These synthetically augmented data are only applicable to our proposed BiDP algorithm. After formulating the script, we finally expanded the data volume by 100 times, which results in 5K$\sim$24K trajectories per task. This magnitude is comparable to existing large-scale bimanual teleoperation methods such as RDT \cite{liu2024rdt} (6K+ self-created episodes) and $\pi_0$ \cite{black2024pi0} (5$\sim$100 hours post-training data), but our cost is extremely low.

\subsubsection{Baselines}
We compare our method to four strong baselines. (1) \textit{Action Chunking Transformers (ACT)} \cite{zhao2023learning}. It is proposed by ALOHA and uses a well-designed transformer structure as the visual encoder. (2) \textit{Diffusion Policy (DP)} \cite{chi2023diffusion}. The vanilla diffusion policy uses RGB images as inputs and ResNet \cite{he2016deep} as the visual encoder. We modified it by using point cloud scenes as observations and a PointNet++ encoder \cite{qi2017pointnet}. (3) \textit{3D Diffusion Policy (DP3)} \cite{ze2024dp3}. It is a variant of diffusion policy with a simpler point cloud encoder. It also designs a two-layer MLP to encode robot proprioceptive states before concatenating with the observation representation. (4) \textit{EquiBot} \cite{yange2024quibot}. It takes the point cloud scene as observation, and learns to predict continuous undecomposed 7-DoF actions of dual arms. Note that these baselines, including our BiDP, are designed to learn task-independent policies, and do not consider the multi-task model currently.


\subsubsection{Metrics}
We train all methods for 500 or 1,000 epochs and only save the last checkpoint for testing. We evaluate each model with 5 trials for each single object (last three tasks with \textbf{30}, \textbf{25} and \textbf{20} trials, respectively) or 2 trials for paired objects (first two tasks with \textbf{54} and \textbf{36} trials, respectively) in every task. These objects have randomized initial placements. For a more detailed comparison, we report the \textbf{average length} (following CLAVIN \cite{mees2022calvin}) in each substep for a sequenced long-horizon task, where the last substep indicates the final \textbf{success rate}. Although above tests have new variations in object placements, we choose two tasks \texttt{pull drawer} and \texttt{uncover lid} to perform more challenging out-of-distribution (OOD) evaluations on novel objects. We omit the last object or paired objects from the training set and treat them as unseen objects to evaluate the final trained model. The number of all OOD trials is quadrupled.

\begin{figure}[t] 
	\begin{center}
           \includegraphics[width=1.0\linewidth]{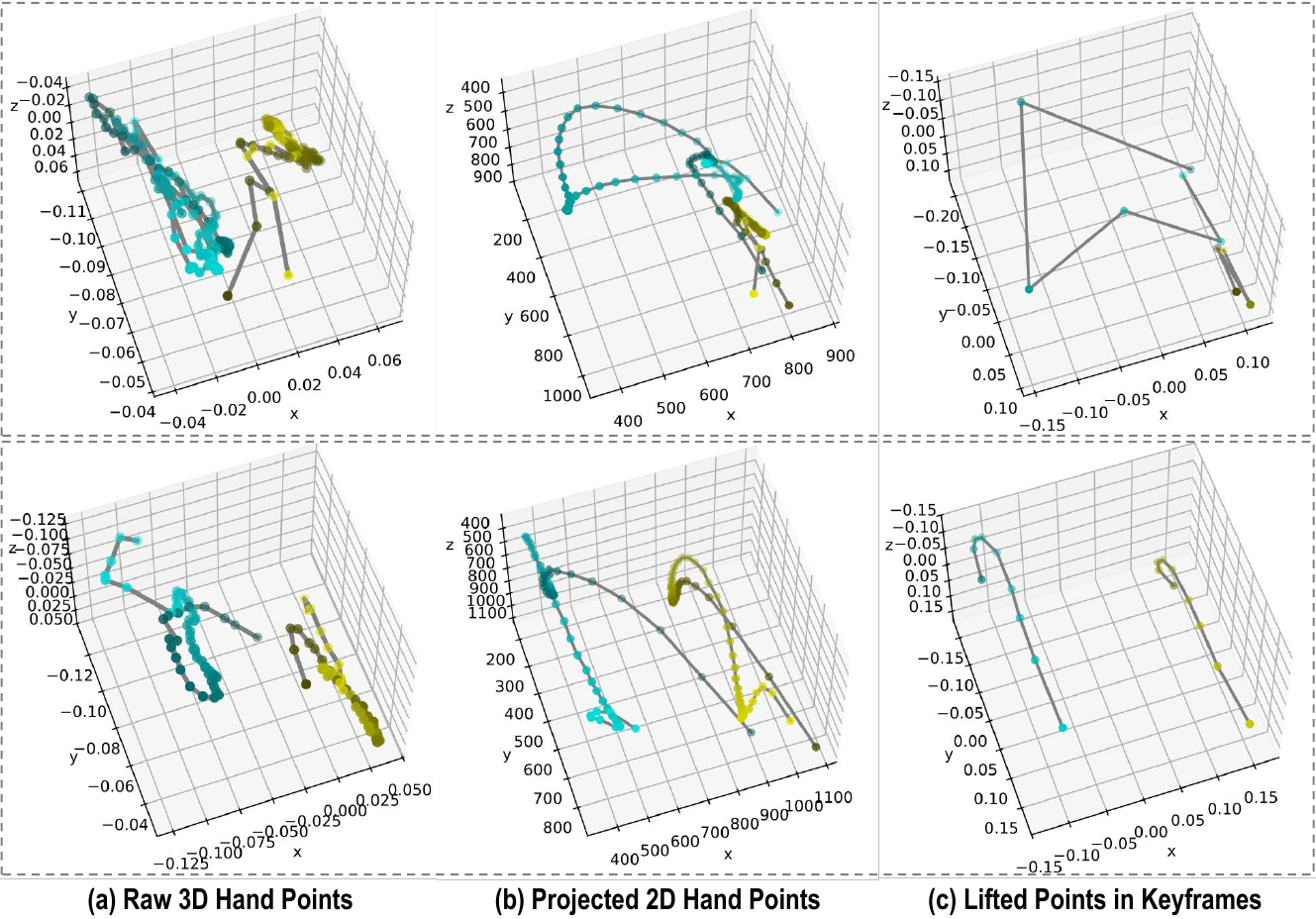}
	\vspace{-10pt}
	\captionof{figure}{Illustrations of extracted hand motion trajectories by using \textbf{(a)} unhandled raw 3D hand center points, \textbf{(b)} projected hand center points on the 2D image, and \textbf{(c)} lifted 3D points in simplified keyframes. The first and second line represents the task \textit{pull drawer} and \textit{uncover lid}, respectively.}
           \label{motions}
	\vspace{-10pt}
	\end{center}
\end{figure}

\setlength{\tabcolsep}{2pt}
\begin{table}[!t]  
	\centering
	\caption{Ablation studies of proposed strategies in YOTO and the bimanual diffusion policy (BiDP). The task \textit{pull drawer} with 243 episodes is used to train all models.}
	\begin{tabular}{c|c|c|c|c|c|c}
	\Xhline{1.2pt}
	\multirow{3}{*}{Ids} & \multirow{3}{*}{\makecell{purely\\object\\observation}} & \multirow{3}{*}{\makecell{using\\sparse\\keyframes}} & \multirow{3}{*}{\makecell{reorganize\\action\\space}} & \multirow{3}{*}{\makecell{using\\geometric\\transforms}} & \multirow{3}{*}{\textbf{\makecell{Success\\Rate}}} & \multirow{3}{*}{\textbf{\makecell{Avg.\\Len.}}} \\
	~ & ~ & ~ & ~ & ~ & ~ & ~ \\
	~ & ~ & ~ & ~ & ~ & ~ & ~ \\
	\Xhline{0.8pt} 
	1 & \xmark & \xmark & \xmark & \xmark & 
		\cellcolor{gray!10}13/54 (24.1\%) & \cellcolor{gray!25}3.54 \\  
	\hline
	2 & \cmark & \xmark & \xmark & \xmark & 
		\cellcolor{gray!10}26/54 (48.1\%) & \cellcolor{gray!25}3.80 \\  
	3 & \xmark & \cmark & \xmark & \xmark & 
		\cellcolor{gray!10}28/54 (51.9\%) & \cellcolor{gray!25}4.15 \\  
	4 & \cmark & \cmark & \xmark & \xmark & 
		\cellcolor{gray!10}31/54 (57.4\%) & \cellcolor{gray!25}4.31 \\  
	5 & \cmark & \cmark & \cmark & \xmark & 
		\cellcolor{gray!10}33/54 (61.1\%) & \cellcolor{gray!25}4.48 \\  
	6 & \cmark & \cmark & \xmark & \cmark & 
		\cellcolor{gray!10}42/54 (77.8\%) & \cellcolor{gray!25}5.15 \\  
	7 & \cmark & \cmark & \cmark & \cmark & 
		\cellcolor{gray!10}43/54 (79.6\%) & \cellcolor{gray!25}5.31 \\  
	\Xhline{1.2pt}
	\end{tabular}
	\label{tabC}
\end{table}

\begin{figure}[!t] 
	\begin{center}
           \includegraphics[width=1.0\linewidth]{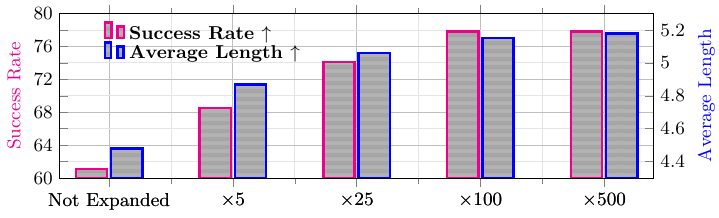}
	\vspace{-15pt}
	\captionof{figure}{Ablation studies on expanded training data at different scales using geometric transformations. The task \textit{pull drawer} with 243 episodes is treated as the not expanded version.}
           \label{scalingBar}
	\vspace{-15pt}
	\end{center}
\end{figure}

\setlength{\tabcolsep}{4pt}
\begin{table*}[!t]  
	\centering
	\caption{Quantitative results of detailed long-horizon performance comparisons (\textbf{in-distribution evaluations}). The step-wise success rates and average length of completed task sequences are reported. We use different colors such as \textcolor{teal}{teal}, \textcolor{olive}{olive} and \textcolor{purple}{purple} to indicate that each substep corresponds to the left arm, right arm and both arms, respectively.}
	\begin{tabular}{c|ccccccc ccccccc c}
	\Xhline{1.2pt}
	\multirow{3}{*}{Methods} & \multicolumn{7}{c|}{pull drawer (243 episodes)} & \multicolumn{7}{c}{pour water (162 episodes)} & ~ \\
	\cline{2-16}
	~ & \textcolor{teal}{\makecell{stable\\drawer}} & \textcolor{olive}{\makecell{pull\\drawer}} & \textcolor{teal}{\makecell{pick\\object}} & \textcolor{teal}{\makecell{place\\object}} & \textcolor{teal}{\makecell{stable\\drawer}} & \textcolor{olive}{\makecell{push\\drawer}} & \multicolumn{1}{c|}{\textbf{\makecell{Avg.\\Len.}}}
	& \textcolor{olive}{\makecell{pick\\mug}} & \textcolor{teal}{\makecell{pick\\bottle}} & \textcolor{olive}{\makecell{close to\\bottle}} & \textcolor{teal}{\makecell{pour\\water}} & \textcolor{teal}{\makecell{place\\bottle}} & \textcolor{olive}{\makecell{place\\mug}} & \textbf{\makecell{Avg.\\Len.}} & ~ \\
	\Xhline{0.8pt} 
	ACT & 42/54 & 26/54 & 18/54 & 15/54 & 09/54 & 05/54 & \multicolumn{1}{c|}{\cellcolor{gray!25}2.13}
		& 28/36 & 24/36 & 23/36 & 03/36 & 03/36 & 03/36 & \cellcolor{gray!25}2.33 \\
	DP & 43/54 & 26/54 & 15/54 & 11/54 & 10/54 & 06/54 & \multicolumn{1}{c|}{\cellcolor{gray!25}2.06} 
		& 30/36 & 29/36 & 29/36 & 06/36 & 06/36 & 06/36 & \cellcolor{gray!25}2.94 \\
	DP3 & 52/54 & 36/54 & 28/54 & 15/54 & 11/54 & 09/54 & \multicolumn{1}{c|}{\cellcolor{gray!25}2.80} 
		& 33/36 & 31/36 & 31/36 & 08/36 & 08/36 & 07/36 & \cellcolor{gray!25}3.28 \\
	EquiBot & 53/54 & 44/54 & 36/54 & 24/54 & 21/54 & 13/54 & \multicolumn{1}{c|}{\cellcolor{gray!25}3.54} 
		& 32/36 & 30/36 & 30/36 & 11/36 & 10/36 & 09/36 & \cellcolor{gray!25}3.39 \\
	BiDP (Ours) & 54/54 & 52/54 & 48/54 & 45/54 & 45/54 & 43/54 & \multicolumn{1}{c|}{\cellcolor{gray!25}5.31} 
		& 35/36 & 34/36 & 34/36 & 29/36 & 28/36 & 28/36 & \cellcolor{gray!25}5.22 \\
	\Xhline{0.8pt} 
	~ & \multicolumn{6}{c|}{unscrew bottle (54 episodes)} & \multicolumn{4}{c|}{uncover lid (45 episodes)} & \multicolumn{5}{c}{open box (36 episodes)} \\
	\cline{2-16}
	~ & \textcolor{teal}{\makecell{pick\\bottle}} & \textcolor{teal}{\makecell{close to\\right}} & \textcolor{olive}{\makecell{unscrew\\cap}} & \textcolor{olive}{\makecell{place\\cap}} & \textcolor{teal}{\makecell{place\\bottle}} & \multicolumn{1}{c|}{\textbf{\makecell{Avg.\\Len.}}} 
	& \textcolor{purple}{\makecell{close\\to lid}} & \textcolor{purple}{\makecell{lift up\\lid}} & \textcolor{purple}{\makecell{place\\lid}} & \multicolumn{1}{c|}{\textbf{\makecell{Avg.\\Len.}}} 
	& \textcolor{purple}{\makecell{close to\\wings}} & \textcolor{purple}{\makecell{open\\wings}} & \textcolor{purple}{\makecell{close to\\wings}} & \textcolor{purple}{\makecell{open\\wings}} & \textbf{\makecell{Avg.\\Len.}} \\
	\Xhline{0.8pt} 
	ACT & 24/30 & 22/30 & 02/30 & 02/30 & 02/30 & \multicolumn{1}{c|}{\cellcolor{gray!25}1.73} 
		& 23/25 & 08/25 & 01/25 & \multicolumn{1}{c|}{\cellcolor{gray!25}1.28} 
		& 15/20 & 05/20 & 05/20 & 00/20 & \cellcolor{gray!25}1.25 \\
	DP & 26/30 & 26/30 & 06/30 & 06/30 & 06/30 & \multicolumn{1}{c|}{\cellcolor{gray!25}2.33} 
		& 23/25 & 16/25 & 04/25 & \multicolumn{1}{c|}{\cellcolor{gray!25}1.72} 
		& 19/20 & 07/20 & 06/20 & 03/20 & \cellcolor{gray!25}1.75 \\
	DP3 & 27/30 & 27/30 & 06/30 & 06/30 & 05/30 & \multicolumn{1}{c|}{\cellcolor{gray!25}2.37} 
		& 24/25 & 19/25 & 06/25 & \multicolumn{1}{c|}{\cellcolor{gray!25}1.96} 
		& 20/20 & 08/20 & 08/20 & 04/20 & \cellcolor{gray!25}2.00 \\
	EquiBot & 28/30 & 28/30 & 08/30 & 07/30 & 06/30 & \multicolumn{1}{c|}{\cellcolor{gray!25}2.57} 
		& 24/25 & 18/25 & 07/25 & \multicolumn{1}{c|}{\cellcolor{gray!25}1.96} 
		& 20/20 & 10/20 & 09/20 & 04/20 & \cellcolor{gray!25}2.35 \\
	BiDP (Ours) & 30/30 & 30/30 & 24/30 & 24/30 & 23/30 & \multicolumn{1}{c|}{\cellcolor{gray!25}4.37} 
		& 25/25 & 24/25 & 20/25 & \multicolumn{1}{c|}{\cellcolor{gray!25}2.76} 
		& 20/20 & 19/20 & 19/20 & 14/20 & \cellcolor{gray!25}3.60 \\
	\Xhline{1.2pt}
	\end{tabular}
	\label{tabD}
	\vspace{-10pt}
\end{table*}

\setlength{\tabcolsep}{5pt}
\begin{table}[!t]  
	\centering
	\caption{Comparison of the average success rate of various methods on all five tasks (\textbf{in-distribution evaluations}).}
	\begin{tabular}{c|ccccc}
	\Xhline{1.2pt}
	Methods & ACT & DP & DP3 & EquiBot & BiDP (Ours) \\
	\Xhline{0.8pt}
	\makecell{Average \\Success Rate} & 5.7\% & 15.8\% & 19.4\% & 23.4\% & 76.8\% \\
	\Xhline{1.2pt}
	\end{tabular}
	\label{tabE}
	\vspace{-10pt}
\end{table}

\setlength{\tabcolsep}{1pt}
\begin{table}[!t]  
	\centering
	\caption{Quantitative results of detailed long-horizon performance comparisons (\textbf{out-of-distribution evaluations}). The substeps are abbreviated as sequential numbers.}
	\begin{tabular}{c|ccccccc cccc | c}
	\Xhline{1.2pt}
	\multirow{3}{*}{Methods} & \multicolumn{7}{c|}{\makecell{pull drawer\\(144 episodes)}} & \multicolumn{4}{c|}{\makecell{uncover lid\\(36 episodes)}} & \multirow{2}{*}{\makecell{Average\\Success\\Rate}} \\
	\cline{2-12}
	~ & \textcolor{teal}{S1} & \textcolor{olive}{S2} & \textcolor{teal}{S3} & \textcolor{teal}{S4} & \textcolor{teal}{S5} & \textcolor{olive}{S6} & \multicolumn{1}{c|}{\textbf{\makecell{Avg.\\Len.}}}
	& \textcolor{purple}{S1} & \textcolor{purple}{S2} & \textcolor{purple}{S3} & \textbf{\makecell{Avg.\\Len.}} & ~ \\
	\Xhline{0.8pt} 
	ACT & 2/8 & 0/8 & 0/8 & 0/8 & 0/8 & 0/8 & \multicolumn{1}{c|}{\cellcolor{gray!25}0.25} 
		& 12/20 & 00/20 & 00/20 & \cellcolor{gray!25}0.60 & \cellcolor{gray!10}0.0\% \\
	DP & 5/8 & 1/8 & 0/8 & 0/8 & 0/8 & 0/8 & \multicolumn{1}{c|}{\cellcolor{gray!25}0.75} 
		& 14/20 & 01/20 & 00/20 & \cellcolor{gray!25}0.75 & \cellcolor{gray!10}0.0\% \\
	DP3 & 5/8 & 1/8 & 1/8 & 0/8 & 0/8 & 0/8 & \multicolumn{1}{c|}{\cellcolor{gray!25}0.88} 
		& 15/20 & 02/20 & 00/20 & \cellcolor{gray!25}0.85 & \cellcolor{gray!10}0.0\% \\
	EquiBot & 5/8 & 3/8 & 3/8 & 3/8 & 3/8 & 1/8 & \multicolumn{1}{c|}{\cellcolor{gray!25}2.25} 
		& 17/20 & 09/20 & 01/20 & \cellcolor{gray!25}1.25 & \cellcolor{gray!10}8.8\% \\
	BiDP (Ours) & 8/8 & 6/8 & 6/8 & 5/8 & 5/8 & 4/8 & \multicolumn{1}{c|}{\cellcolor{gray!25}4.25} 
		& 18/20 & 12/20 & 04/20 & \cellcolor{gray!25}1.70 & \cellcolor{gray!10}35.0\% \\
	\Xhline{1.2pt}
	\end{tabular}
	\label{tabF}
	\vspace{-10pt}
\end{table}

\begin{figure*}[t] 
	\begin{center}
           \includegraphics[width=1.0\linewidth]{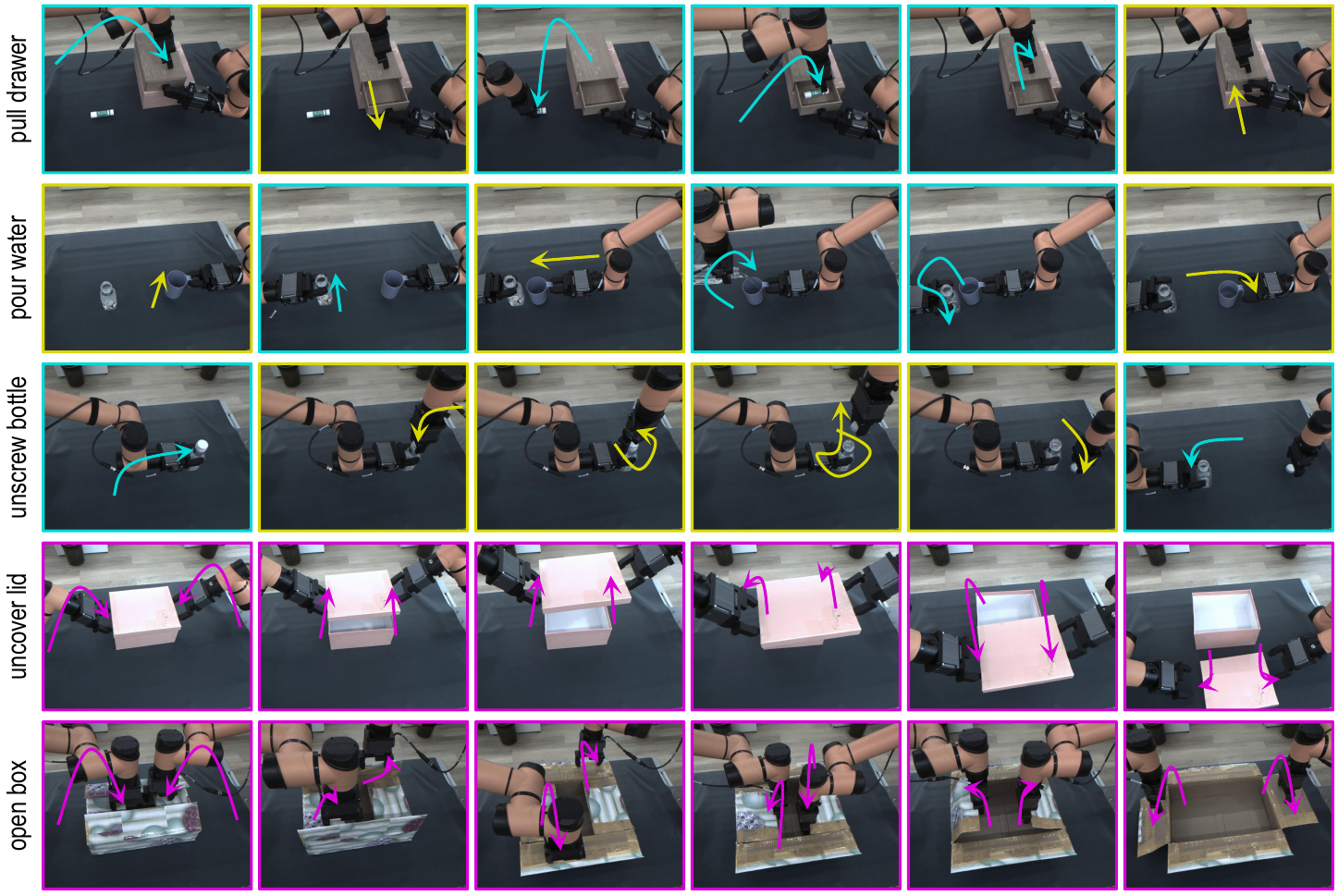}
	\vspace{-15pt}  
	\captionof{figure}{Visualization of five bimanual tasks performed on real robots. We use different colors such as \textcolor{teal}{teal}, \textcolor{olive}{olive} and \textcolor{purple}{purple} to distinguish frames of left arm, right arm and both arms, respectively. Arrows are artificially added to show movement trends.}
           \label{visresults}
	\vspace{-15pt}
	\end{center}
\end{figure*}

\subsection{Results Comparison}
Here, we answer the questions raised at the beginning one by one, including basic in-distribution results and generalizations to out-of-distribution settings.

($\mathsf{Q}1$) \textbf{Our extracted hand motions have good continuity and consistency}.
We first discuss the quality of the extracted motion trajectories, which is the core concept of this paper and extremely important for the various strategies developed next. As shown in Fig.~\ref{motions}, we compared the general effect of 3D hand motion trajectories extracted using different methods in two different long-horizon bimanual tasks. Firstly, when directly applying advanced 3D hand mesh reconstruction methods (either HaMeR \cite{pavlakos2024reconstructing} or WiLoR \cite{potamias2024wilor}), the resulting hand trajectory is always unstable and difficult to parse (see Fig.~\ref{motions} \textbf{(a)}). This is mainly because most of these methods are based on monocular images, and the preset camera parameters such as focus and focal length are directly calculated using the center and size of each image. This makes the estimation results for consecutive frames in the video not in a unified and invariant camera space, and therefore unreliable and ambiguous in depth. Nevertheless, this intuitive but sub-optimal approach is still widely used by mainstream methods for learning from human videos \cite{kerr2024robot, li2024okami, fu2024humanplus}. In comparison, after projecting these 3D points onto a 2D image plane (with the Z-axis set to 0 for ease of visualization), it is clear that the trajectory trends and estimated motion flow are improved (see Fig.~\ref{motions} \textbf{(b)}). This conclusion is generally applicable, for tasks like ours where the camera is stationary and its intrinsic and extrinsic parameters are known. Finally, as described in Sec~\ref{mHMEI}, we filter out sparse keyframes from these continuous points and lift the corresponding position components into 3D points to obtain the keyposes suitable for the end-effector (see Fig.~\ref{motions} \textbf{(c)}). We thus claim that our extracted hand motion trajectory based on an one-shot human teaching has a more guaranteed quality. And we expect that this motion extraction technology will be used for retargeting to other more dexterous end-effectors, such as multi-fingered hands. 

($\mathsf{Q}2$) \textbf{The various strategies we propose in YOTO are effective}.
After extracting primary keyposes that could be successfully injected into the robot, we continue to explore YOTO including other strategies, which are closely related to the visuomotor policy learning. As shown in Tab.~\ref{tabC}, we quantitatively illustrate the effectiveness of each strategy one by one through many ablation studies. We experimented with task \texttt{pull drawer} which has 243 training trajectories. First, the method (\textit{id-1}) without any proposed strategy can be regarded as the vanilla EquiBot \cite{yange2024quibot}, which takes the entire point cloud scene as observation, learns to predict continuous actions, models paired end-effector poses and leverages non-augmented training demonstrations. Despite being a solid baseline, it performed the worst on this challenging long-horizon task. Next, we replaced the input with point clouds containing only manipulated objects (\textit{id-2}) or predicted simplified sparse keyposes (\textit{id-3}), and the success rate and average execution length of the task were improved. These results suggest that reducing unnecessary distractions in the input and learning fewer simplified actions are the right direction. When both are used together (\textit{id-4}), better performance can be achieved. Based on these two strategies, we decoupled the output action space and reconstructed it into a single-arm format (\textit{id-5}), the policy could also be superior, indicating the importance of eliminating redundant actions. Alternately, if 3D geometric transformations were applied to further expand training demonstrations (\textit{id-6}), the resulting model effect was much better, with the most prominent growth. This proves that our developed demonstration proliferation is simple yet efficient. We accordingly show in Fig.~\ref{scalingBar} the typical trend that using more extended training data leads to better performance, which is consistent with our consensus. Finally, combining the above strategies together (\textit{id-7}), our BiDP takes full advantage of all the strengths and has achieved the best results.

On the other hand, we need to compare and explain whether BiDP is better than other visuomotor imitation methods \cite{zhao2023learning, chi2023diffusion, ze2024dp3, yange2024quibot} on more bimanual tasks. As shown in Tab.~\ref{tabD}, following the mainstream in-distribution setting, we performed extensive policies training and real robot evaluations on five long-horizon tasks, and reported a detailed performance comparison of various methods. Generally speaking, we can draw three conclusions from these quantitative data. (1) First, the diffusion-based strategy always performed better than the transformer-based ACT. This is mainly because the diffusion model can model a higher-dimensional action space and is highly malleable, while tranformer architectures usually do not have these characteristics and require a large amount of data to achieve scale effects and gain advantages. In addition, ACT utilizes 2D images as observations instead of 3D input, which also makes it achieve inferior results. (2) Second, a more advanced and sophisticated 3D observation perception architecture can lead to higher policy performance. For example, compared to the modified DP that directly uses PointNet++ to process 3D point cloud input, DP3 and EquiBot adopt a self-designed lightweight MLP encoder and SIM(3)-equivariant backbone to extract point cloud features, respectively, and always achieved better results. (3) Finally, for more complex long-horizon bimanual manipulation tasks, the existing state-of-the-art methods still have a lot of room for improvement, such as the gradually decaying effect over multiple substeps and less exploration of efficient utilization of training data. Thanks to the proposed multiple strategies, our BiDP can better cope with bimanual tasks, significantly better than all compared policies. We summarized the average success rate of each method on all five tasks in Tab~\ref{tabE}, where our method BiDP achieved a success rate of nearly 60\%, demonstrating good potential for practical robotic applications. To sum up, it can be concluded that the various strategies we proposed in YOTO are quite effective.

\begin{figure*}[] 
	\begin{center}
           \includegraphics[width=1.0\linewidth]{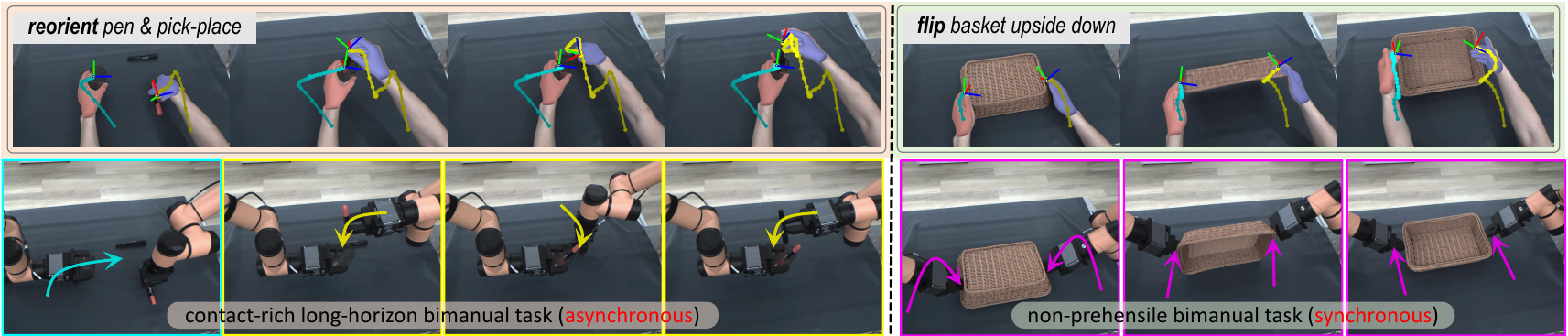}
	\vspace{-10pt}  
	\captionof{figure}{Illustrations of another two bimanual tasks. \textbf{Top:} the visualization of hand motions extraction. \textbf{Bottom:} the corresponding rollout examples by injecting actions on real robots. Refer to Fig.~\ref{teaser} and Fig.~\ref{visresults} for notes on different colors and curves.}
           \label{newtasks}
	\vspace{-10pt}
	\end{center}
\end{figure*}

($\mathsf{Q}3$) \textbf{BiDP has satisfactory out-of-domain generalization ability}.
To further illustrate the superiority of BiDP, we designed tests under out-of-distribution (OOD) settings. Results are shown in Tab~\ref{tabF}. From it, we can see that, except for our method and EquiBot, the performance of the other three methods has dropped significantly when it comes to OOD setups, showing poor generalization to unseen objects. Comparing to EquiBot, our BiDP still has a clear advantage, thanks to the fact that we use explicit 3D geometric transformations for expanding the training demonstrations instead of SIM(3)-equivariant augmentation of the entire point cloud input in EquiBot. In addition, using pure object point clouds as input also makes our model more robust compared to all baselines. The core idea here is to rely on the still rapidly developing capabilities of vision foundation models, such as the open vocabulary detection \cite{xiao2024florence} and segmentation \cite{ravi2024sam}, to more reliably  perceive various unseen scenes and objects. In summary, these results verify that our BiDP indeed outperforms prior methods with the least amount of performance degradation in OOD generalization.

($\mathsf{Q}4$) \textbf{YOTO is widely applicable to diverse bimanual tasks}.
Our proposed framework YOTO is compatible with most bimanual tasks, such as the selected five representative long-horizon tasks, covering a variety of skills, multi-object perception, dual-arm coordinated processing, intricate motion trajectories, and varying execution substeps. In addition to the above-mentioned quantitative results, we also qualitatively demonstrate the visual effects of real robot execution on five tasks in Fig.~\ref{visresults}, mainly showing the sparse keyframes contained in them. We can see that the two robot arms have learned the movements demonstrated by human hands and complete these complex tasks in an orderly manner. 

Moreover, we selected another two typical bimanual manipulation tasks and enabled the dual-arm robot to learn new given tasks quickly and easily through one-shot human teaching. Due to space limitations, we did not continue the demonstration proliferation and policy training. The illustrations of extracted actions that can be injected into real robots are shown in Fig.~\ref{newtasks}. These results further reveal the simplicity, versatility and scalability of YOTO. In the future, we will explore using YOTO to handle more intricate, valuable, but less researched bimanual manipulation tasks.


\section{Conclusion and Limitation}


In this paper, we propose a novel framework named YOTO to address the challenge of efficient and robust bimanual manipulation. Our approach learns from one-shot human video demonstrations, using vision techniques to extract fine-grained and consecutive hand features like pose, joints, and states. To ensure stable and precise manipulation, we simplify noisy hand motion trajectories into discrete keyframes and introduce a motion mask for better dual-arm coordination. Based on the refined one-shot teaching, we develop a scalable data proliferation solution using auto-rollout verification and 3D geometric transformations to rapidly create diverse training examples. With this enriched dataset, we design a dedicated bimanual diffusion policy (BiDP) that simplifies observations, predicts keyposes, and reorganizes action spaces for efficient training. Validated on five complex bimanual tasks, our framework demonstrates superior performance in both synchronous and asynchronous scenarios. These contributions provide a standardized method for transferring human motions to robots, a scalable approach for data generation, and an effective algorithm for mastering intricate dual-arm tasks, advancing the field of bimanual manipulation.

\textbf{Limitation:} Although YOTO has achieved impressive performance on various long-horizon bimanual manipulation tasks, we conclude that it has at least the following limitations. (1) Our vision-based hand trajectory extraction schemes have inherent errors. This means that we have to check carefully and verify on the real robot whether the extracted position and posture information is reliable, which still requires additional manpower. (2) The primary version of YOTO adopts a fixed workbench, which limits its flexibility and accessibility. In the future, we may consider using mobile bases, such as wheeled carts or multi-legged robots. (3) The equipped parallel gripper is not flexible enough and has limited functionality. Upgrading the end-effector to a multi-fingered dexterous hand or equipping it with force-tactile sensors can make the robot more versatile and powerful. (4) More ultra-difficult bimanual tasks are still under-explored, such as the specialized tool-based manipulation (\textit{e.g.}, picking up a hammer to pound a nail or twisting a screwdriver to tighten a screw), highly dynamic non-quasi-stationary tasks, and friendly interactive collaboration with people. In short, these limitations highlight the need for further innovations to enhance robustness, generalization, and scalability in bimanual robot manipulation.


\section*{Acknowledgments}


This work was supported by the Guangdong Provincial Key Field R\&D Program (No. 20240104, the project name: Research and Application of Common Key Technologies of Robot Embodied Intelligence Based on AI Large Model), and also received funding from the 2024 Shenzhen Science and Technology Major Project (No. 202402002, the project name: Research and Development of Multimodal Database for Robots to Learn Human Skills).

\bibliographystyle{plainnat}
\bibliography{refs}

\begin{thebibliography}{108}
\providecommand{\natexlab}[1]{#1}
\providecommand{\url}[1]{\texttt{#1}}
\expandafter\ifx\csname urlstyle\endcsname\relax
  \providecommand{\doi}[1]{doi: #1}\else
  \providecommand{\doi}{doi: \begingroup \urlstyle{rm}\Url}\fi

\bibitem[Aldaco et~al.(2024)Aldaco, Armstrong, Baruch, Bingham, Chan, Draper,
  Dwibedi, Finn, Florence, Goodrich, et~al.]{aldaco2024aloha}
Jorge Aldaco, Travis Armstrong, Robert Baruch, Jeff Bingham, Sanky Chan,
  Kenneth Draper, Debidatta Dwibedi, Chelsea Finn, Pete Florence, Spencer
  Goodrich, et~al.
\newblock Aloha 2: An enhanced low-cost hardware for bimanual teleoperation.
\newblock \emph{arXiv preprint arXiv:2405.02292}, 2024.

\bibitem[Avigal et~al.(2022)Avigal, Berscheid, Asfour, Kr{\"o}ger, and
  Goldberg]{avigal2022speedfolding}
Yahav Avigal, Lars Berscheid, Tamim Asfour, Torsten Kr{\"o}ger, and Ken
  Goldberg.
\newblock Speedfolding: Learning efficient bimanual folding of garments.
\newblock In \emph{2022 IEEE/RSJ International Conference on Intelligent Robots
  and Systems (IROS)}, pages 1--8. IEEE, 2022.

\bibitem[Bahety et~al.(2023)Bahety, Jain, Ha, Hager, Burchfiel, Cousineau,
  Feng, and Song]{bahety2023bag}
Arpit Bahety, Shreeya Jain, Huy Ha, Nathalie Hager, Benjamin Burchfiel, Eric
  Cousineau, Siyuan Feng, and Shuran Song.
\newblock Bag all you need: Learning a generalizable bagging strategy for
  heterogeneous objects.
\newblock In \emph{2023 IEEE/RSJ International Conference on Intelligent Robots
  and Systems (IROS)}, pages 960--967. IEEE, 2023.

\bibitem[Bahety et~al.(2024)Bahety, Mandikal, Abbatematteo, and
  Mart{\'\i}n-Mart{\'\i}n]{bahety2024screwmimic}
Arpit Bahety, Priyanka Mandikal, Ben Abbatematteo, and Roberto
  Mart{\'\i}n-Mart{\'\i}n.
\newblock Screwmimic: Bimanual imitation from human videos with screw space
  projection.
\newblock In \emph{Proceedings of Robotics: Science and Systems (RSS)}, 2024.

\bibitem[Balakrishnan and Kurtenbach(1999)]{balakrishnan1999exploring}
Ravin Balakrishnan and Gordon Kurtenbach.
\newblock Exploring bimanual camera control and object manipulation in 3d
  graphics interfaces.
\newblock In \emph{Proceedings of the SIGCHI Conference on Human Factors in
  Computing Systems}, pages 56--62, 1999.

\bibitem[Bersch et~al.(2011)Bersch, Pitzer, and Kammel]{bersch2011bimanual}
Christian Bersch, Benjamin Pitzer, and S{\"o}ren Kammel.
\newblock Bimanual robotic cloth manipulation for laundry folding.
\newblock In \emph{2011 IEEE/RSJ International Conference on Intelligent Robots
  and Systems}, pages 1413--1419. IEEE, 2011.

\bibitem[Black et~al.(2024)Black, Brown, Driess, Esmail, Equi, Finn, Fusai,
  Groom, Hausman, Ichter, et~al.]{black2024pi0}
Kevin Black, Noah Brown, Danny Driess, Adnan Esmail, Michael Equi, Chelsea
  Finn, Niccolo Fusai, Lachy Groom, Karol Hausman, Brian Ichter, et~al.
\newblock $\pi_0$: A vision-language-action flow model for general robot
  control.
\newblock \emph{arXiv preprint arXiv:2410.24164}, 2024.

\bibitem[Brehmer et~al.(2024)Brehmer, Bose, De~Haan, and
  Cohen]{brehmer2024edgi}
Johann Brehmer, Joey Bose, Pim De~Haan, and Taco~S Cohen.
\newblock Edgi: Equivariant diffusion for planning with embodied agents.
\newblock \emph{Advances in Neural Information Processing Systems}, 36, 2024.

\bibitem[Canberk et~al.(2023)Canberk, Chi, Ha, Burchfiel, Cousineau, Feng, and
  Song]{canberk2023cloth}
Alper Canberk, Cheng Chi, Huy Ha, Benjamin Burchfiel, Eric Cousineau, Siyuan
  Feng, and Shuran Song.
\newblock Cloth funnels: Canonicalized-alignment for multi-purpose garment
  manipulation.
\newblock In \emph{2023 IEEE International Conference on Robotics and
  Automation (ICRA)}, pages 5872--5879. IEEE, 2023.

\bibitem[Chen et~al.(2023{\natexlab{a}})Chen, Shi, Lin, Seita, Ahmad, Cheng,
  Kollar, Held, and Goldberg]{chen2023bagging}
Lawrence~Yunliang Chen, Baiyu Shi, Roy Lin, Daniel Seita, Ayah Ahmad, Richard
  Cheng, Thomas Kollar, David Held, and Ken Goldberg.
\newblock Bagging by learning to singulate layers using interactive perception.
\newblock In \emph{2023 IEEE/RSJ International Conference on Intelligent Robots
  and Systems (IROS)}, pages 3176--3183. IEEE, 2023{\natexlab{a}}.

\bibitem[Chen et~al.(2023{\natexlab{b}})Chen, Shi, Seita, Cheng, Kollar, Held,
  and Goldberg]{chen2023autobag}
Lawrence~Yunliang Chen, Baiyu Shi, Daniel Seita, Richard Cheng, Thomas Kollar,
  David Held, and Ken Goldberg.
\newblock Autobag: Learning to open plastic bags and insert objects.
\newblock In \emph{2023 IEEE International Conference on Robotics and
  Automation (ICRA)}, pages 3918--3925. IEEE, 2023{\natexlab{b}}.

\bibitem[Chen et~al.(2024{\natexlab{a}})Chen, Wang, Nguyen, Fei-Fei, and
  Liu]{chen2024arcap}
Sirui Chen, Chen Wang, Kaden Nguyen, Li~Fei-Fei, and C~Karen Liu.
\newblock Arcap: Collecting high-quality human demonstrations for robot
  learning with augmented reality feedback.
\newblock \emph{arXiv preprint arXiv:2410.08464}, 2024{\natexlab{a}}.

\bibitem[Chen et~al.(2024{\natexlab{b}})Chen, Wang, Yang, and
  Liu]{chen2024object}
Yuanpei Chen, Chen Wang, Yaodong Yang, and Karen Liu.
\newblock Object-centric dexterous manipulation from human motion data.
\newblock In \emph{8th Annual Conference on Robot Learning},
  2024{\natexlab{b}}.

\bibitem[Cheng et~al.(2024)Cheng, Li, Yang, Yang, and Wang]{cheng2024open}
Xuxin Cheng, Jialong Li, Shiqi Yang, Ge~Yang, and Xiaolong Wang.
\newblock Open-television: Teleoperation with immersive active visual feedback.
\newblock In \emph{8th Annual Conference on Robot Learning}, 2024.

\bibitem[Chi et~al.(2023)Chi, Xu, Feng, Cousineau, Du, Burchfiel, Tedrake, and
  Song]{chi2023diffusion}
Cheng Chi, Zhenjia Xu, Siyuan Feng, Eric Cousineau, Yilun Du, Benjamin
  Burchfiel, Russ Tedrake, and Shuran Song.
\newblock Diffusion policy: Visuomotor policy learning via action diffusion.
\newblock \emph{The International Journal of Robotics Research}, page
  02783649241273668, 2023.

\bibitem[Chitnis et~al.(2020)Chitnis, Tulsiani, Gupta, and
  Gupta]{chitnis2020efficient}
Rohan Chitnis, Shubham Tulsiani, Saurabh Gupta, and Abhinav Gupta.
\newblock Efficient bimanual manipulation using learned task schemas.
\newblock In \emph{2020 IEEE International Conference on Robotics and
  Automation (ICRA)}, pages 1149--1155. IEEE, 2020.

\bibitem[Chuang et~al.(2024)Chuang, Lee, Gao, and Soltani]{chuang2024active}
Ian Chuang, Andrew Lee, Dechen Gao, and Iman Soltani.
\newblock Active vision might be all you need: Exploring active vision in
  bimanual robotic manipulation.
\newblock \emph{arXiv preprint arXiv:2409.17435}, 2024.

\bibitem[Ciocarlie et~al.(2007)Ciocarlie, Goldfeder, and
  Allen]{ciocarlie2007dexterous}
Matei Ciocarlie, Corey Goldfeder, and Peter Allen.
\newblock Dexterous grasping via eigengrasps: A low-dimensional approach to a
  high-complexity problem.
\newblock In \emph{Proceedings of Robotics: Science and Systems (RSS)}, 2007.

\bibitem[Colom{\'e} and Torras(2018)]{colome2018dimensionality}
Adria Colom{\'e} and Carme Torras.
\newblock Dimensionality reduction for dynamic movement primitives and
  application to bimanual manipulation of clothes.
\newblock \emph{IEEE Transactions on Robotics}, 34\penalty0 (3):\penalty0
  602--615, 2018.

\bibitem[Ding et~al.(2024)Ding, Qin, Zhu, Jia, Yang, Yang, Qi, and
  Wang]{ding2024bunny}
Runyu Ding, Yuzhe Qin, Jiyue Zhu, Chengzhe Jia, Shiqi Yang, Ruihan Yang,
  Xiaojuan Qi, and Xiaolong Wang.
\newblock Bunny-visionpro: Real-time bimanual dexterous teleoperation for
  imitation learning.
\newblock \emph{arXiv preprint arXiv:2407.03162}, 2024.

\bibitem[Drolet et~al.(2024)Drolet, Stepputtis, Kailas, Jain, Peters, Schaal,
  and Amor]{drolet2024comparison}
Michael Drolet, Simon Stepputtis, Siva Kailas, Ajinkya Jain, Jan Peters, Stefan
  Schaal, and Heni~Ben Amor.
\newblock A comparison of imitation learning algorithms for bimanual
  manipulation.
\newblock \emph{IEEE Robotics and Automation Letters}, 2024.

\bibitem[Fan et~al.(2023)Fan, Taheri, Tzionas, Kocabas, Kaufmann, Black, and
  Hilliges]{fan2023arctic}
Zicong Fan, Omid Taheri, Dimitrios Tzionas, Muhammed Kocabas, Manuel Kaufmann,
  Michael~J Black, and Otmar Hilliges.
\newblock Arctic: A dataset for dexterous bimanual hand-object manipulation.
\newblock In \emph{Proceedings of the IEEE/CVF Conference on Computer Vision
  and Pattern Recognition}, pages 12943--12954, 2023.

\bibitem[Fu et~al.(2024{\natexlab{a}})Fu, Zhao, Wu, Wetzstein, and
  Finn]{fu2024humanplus}
Zipeng Fu, Qingqing Zhao, Qi~Wu, Gordon Wetzstein, and Chelsea Finn.
\newblock Humanplus: Humanoid shadowing and imitation from humans.
\newblock \emph{arXiv preprint arXiv:2406.10454}, 2024{\natexlab{a}}.

\bibitem[Fu et~al.(2024{\natexlab{b}})Fu, Zhao, and Finn]{fu2024mobile}
Zipeng Fu, Tony~Z Zhao, and Chelsea Finn.
\newblock Mobile aloha: Learning bimanual mobile manipulation with low-cost
  whole-body teleoperation.
\newblock \emph{arXiv preprint arXiv:2401.02117}, 2024{\natexlab{b}}.

\bibitem[Gao et~al.(2024)Gao, Jin, Krebs, Jaquier, and Asfour]{gao2024bi}
Jianfeng Gao, Xiaoshu Jin, Franziska Krebs, No{\'e}mie Jaquier, and Tamim
  Asfour.
\newblock Bi-kvil: Keypoints-based visual imitation learning of bimanual
  manipulation tasks.
\newblock In \emph{2024 IEEE International Conference on Robotics and
  Automation (ICRA)}, pages 16850--16857. IEEE, 2024.

\bibitem[Goyal et~al.(2024)Goyal, Blukis, Xu, Guo, Chao, and Fox]{goyal2024rvt}
Ankit Goyal, Valts Blukis, Jie Xu, Yijie Guo, Yu-Wei Chao, and Dieter Fox.
\newblock Rvt-2: Learning precise manipulation from few demonstrations.
\newblock In \emph{Proceedings of Robotics: Science and Systems (RSS)}, 2024.

\bibitem[Grannen et~al.(2021)Grannen, Sundaresan, Thananjeyan, Ichnowski,
  Balakrishna, Viswanath, Laskey, Gonzalez, and
  Goldberg]{grannen2021untangling}
Jennifer Grannen, Priya Sundaresan, Brijen Thananjeyan, Jeffrey Ichnowski,
  Ashwin Balakrishna, Vainavi Viswanath, Michael Laskey, Joseph Gonzalez, and
  Ken Goldberg.
\newblock Untangling dense knots by learning task-relevant keypoints.
\newblock In \emph{Conference on Robot Learning}, pages 782--800. PMLR, 2021.

\bibitem[Grannen et~al.(2023{\natexlab{a}})Grannen, Wu, Belkhale, and
  Sadigh]{grannen2023learning}
Jennifer Grannen, Yilin Wu, Suneel Belkhale, and Dorsa Sadigh.
\newblock Learning bimanual scooping policies for food acquisition.
\newblock In \emph{Conference on Robot Learning}, pages 1510--1519. PMLR,
  2023{\natexlab{a}}.

\bibitem[Grannen et~al.(2023{\natexlab{b}})Grannen, Wu, Vu, and
  Sadigh]{grannen2023stabilize}
Jennifer Grannen, Yilin Wu, Brandon Vu, and Dorsa Sadigh.
\newblock Stabilize to act: Learning to coordinate for bimanual manipulation.
\newblock In \emph{Conference on Robot Learning}, pages 563--576. PMLR,
  2023{\natexlab{b}}.

\bibitem[Grauman et~al.(2022)Grauman, Westbury, Byrne, Chavis, Furnari,
  Girdhar, Hamburger, Jiang, Liu, Liu, et~al.]{grauman2022ego4d}
Kristen Grauman, Andrew Westbury, Eugene Byrne, Zachary Chavis, Antonino
  Furnari, Rohit Girdhar, Jackson Hamburger, Hao Jiang, Miao Liu, Xingyu Liu,
  et~al.
\newblock Ego4d: Around the world in 3,000 hours of egocentric video.
\newblock In \emph{Proceedings of the IEEE/CVF Conference on Computer Vision
  and Pattern Recognition}, pages 18995--19012, 2022.

\bibitem[Grauman et~al.(2024)Grauman, Westbury, Torresani, Kitani, Malik,
  Afouras, Ashutosh, Baiyya, Bansal, Boote, et~al.]{grauman2024ego}
Kristen Grauman, Andrew Westbury, Lorenzo Torresani, Kris Kitani, Jitendra
  Malik, Triantafyllos Afouras, Kumar Ashutosh, Vijay Baiyya, Siddhant Bansal,
  Bikram Boote, et~al.
\newblock Ego-exo4d: Understanding skilled human activity from first-and
  third-person perspectives.
\newblock In \emph{Proceedings of the IEEE/CVF Conference on Computer Vision
  and Pattern Recognition}, pages 19383--19400, 2024.

\bibitem[Grotz et~al.(2024)Grotz, Shridhar, Asfour, and Fox]{grotz2024peract2}
Markus Grotz, Mohit Shridhar, Tamim Asfour, and Dieter Fox.
\newblock Peract2: Benchmarking and learning for robotic bimanual manipulation
  tasks.
\newblock \emph{arXiv preprint arXiv:2407.00278}, 2024.

\bibitem[Hartmann et~al.(2022)Hartmann, Orthey, Driess, Oguz, and
  Toussaint]{hartmann2022long}
Valentin~N Hartmann, Andreas Orthey, Danny Driess, Ozgur~S Oguz, and Marc
  Toussaint.
\newblock Long-horizon multi-robot rearrangement planning for construction
  assembly.
\newblock \emph{IEEE Transactions on Robotics}, 39\penalty0 (1):\penalty0
  239--252, 2022.

\bibitem[He et~al.(2016)He, Zhang, Ren, and Sun]{he2016deep}
Kaiming He, Xiangyu Zhang, Shaoqing Ren, and Jian Sun.
\newblock Deep residual learning for image recognition.
\newblock In \emph{Proceedings of the IEEE Conference on Computer Vision and
  Pattern Recognition}, pages 770--778, 2016.

\bibitem[Hebert et~al.(2013)Hebert, Hudson, Ma, and Burdick]{hebert2013dual}
Paul Hebert, Nicolas Hudson, Jeremy Ma, and Joel~W Burdick.
\newblock Dual arm estimation for coordinated bimanual manipulation.
\newblock In \emph{2013 IEEE International Conference on Robotics and
  Automation}, pages 120--125. IEEE, 2013.

\bibitem[Ho et~al.(2020)Ho, Jain, and Abbeel]{ho2020denoising}
Jonathan Ho, Ajay Jain, and Pieter Abbeel.
\newblock Denoising diffusion probabilistic models.
\newblock \emph{Advances in Neural Information Processing Systems},
  33:\penalty0 6840--6851, 2020.

\bibitem[Huang et~al.(2023)Huang, Chen, Wang, Qin, Yang, Atanasov, and
  Wang]{huang2023dynamic}
Binghao Huang, Yuanpei Chen, Tianyu Wang, Yuzhe Qin, Yaodong Yang, Nikolay
  Atanasov, and Xiaolong Wang.
\newblock Dynamic handover: Throw and catch with bimanual hands.
\newblock In \emph{Conference on Robot Learning}, pages 1887--1902. PMLR, 2023.

\bibitem[James et~al.(2022)James, Wada, Laidlow, and Davison]{james2022coarse}
Stephen James, Kentaro Wada, Tristan Laidlow, and Andrew~J Davison.
\newblock Coarse-to-fine q-attention: Efficient learning for visual robotic
  manipulation via discretisation.
\newblock In \emph{Proceedings of the IEEE/CVF Conference on Computer Vision
  and Pattern Recognition}, pages 13739--13748, 2022.

\bibitem[Jang et~al.(2022)Jang, Irpan, Khansari, Kappler, Ebert, Lynch, Levine,
  and Finn]{jang2022bc}
Eric Jang, Alex Irpan, Mohi Khansari, Daniel Kappler, Frederik Ebert, Corey
  Lynch, Sergey Levine, and Chelsea Finn.
\newblock Bc-z: Zero-shot task generalization with robotic imitation learning.
\newblock In \emph{Conference on Robot Learning}, pages 991--1002. PMLR, 2022.

\bibitem[Johns(2021)]{johns2021coarse}
Edward Johns.
\newblock Coarse-to-fine imitation learning: Robot manipulation from a single
  demonstration.
\newblock In \emph{2021 IEEE international conference on robotics and
  automation (ICRA)}, pages 4613--4619. IEEE, 2021.

\bibitem[Ke et~al.(2024)Ke, Gkanatsios, and Fragkiadaki]{ke20243d}
Tsung-Wei Ke, Nikolaos Gkanatsios, and Katerina Fragkiadaki.
\newblock 3d diffuser actor: Policy diffusion with 3d scene representations.
\newblock \emph{arXiv preprint arXiv:2402.10885}, 2024.

\bibitem[Kerr et~al.(2024)Kerr, Kim, Wu, Yi, Wang, Goldberg, and
  Kanazawa]{kerr2024robot}
Justin Kerr, Chung~Min Kim, Mingxuan Wu, Brent Yi, Qianqian Wang, Ken Goldberg,
  and Angjoo Kanazawa.
\newblock Robot see robot do: Imitating articulated object manipulation with
  monocular 4d reconstruction.
\newblock In \emph{8th Annual Conference on Robot Learning}, 2024.

\bibitem[Kim et~al.(2024)Kim, Pertsch, Karamcheti, Xiao, Balakrishna, Nair,
  Rafailov, Foster, Lam, Sanketi, et~al.]{kim2024openvla}
Moo~Jin Kim, Karl Pertsch, Siddharth Karamcheti, Ted Xiao, Ashwin Balakrishna,
  Suraj Nair, Rafael Rafailov, Ethan Foster, Grace Lam, Pannag Sanketi, et~al.
\newblock Openvla: An open-source vision-language-action model.
\newblock \emph{arXiv preprint arXiv:2406.09246}, 2024.

\bibitem[Ko et~al.(2024)Ko, Mao, Du, Sun, and Tenenbaum]{ko2024learning}
Po-Chen Ko, Jiayuan Mao, Yilun Du, Shao-Hua Sun, and Joshua~B Tenenbaum.
\newblock Learning to act from actionless videos through dense correspondences.
\newblock In \emph{The Twelfth International Conference on Learning
  Representations}, 2024.

\bibitem[Krebs and Asfour(2022)]{krebs2022bimanual}
Franziska Krebs and Tamim Asfour.
\newblock A bimanual manipulation taxonomy.
\newblock \emph{IEEE Robotics and Automation Letters}, 7\penalty0 (4):\penalty0
  11031--11038, 2022.

\bibitem[Li et~al.(2024{\natexlab{a}})Li, Tsagkas, Song, Mon-Williams,
  Vijayakumar, Shao, and Sevilla-Lara]{li2024learning}
Gen Li, Nikolaos Tsagkas, Jifei Song, Ruaridh Mon-Williams, Sethu Vijayakumar,
  Kun Shao, and Laura Sevilla-Lara.
\newblock Learning precise affordances from egocentric videos for robotic
  manipulation.
\newblock \emph{arXiv preprint arXiv:2408.10123}, 2024{\natexlab{a}}.

\bibitem[Li et~al.(2024{\natexlab{b}})Li, Zhu, Xie, Jiang, Seo, Pavlakos, and
  Zhu]{li2024okami}
Jinhan Li, Yifeng Zhu, Yuqi Xie, Zhenyu Jiang, Mingyo Seo, Georgios Pavlakos,
  and Yuke Zhu.
\newblock Okami: Teaching humanoid robots manipulation skills through single
  video imitation.
\newblock In \emph{8th Annual Conference on Robot Learning},
  2024{\natexlab{b}}.

\bibitem[Li et~al.(2023)Li, Pan, Xu, Wang, and Wu]{li2023efficient}
Yunfei Li, Chaoyi Pan, Huazhe Xu, Xiaolong Wang, and Yi~Wu.
\newblock Efficient bimanual handover and rearrangement via symmetry-aware
  actor-critic learning.
\newblock In \emph{2023 IEEE International Conference on Robotics and
  Automation (ICRA)}, pages 3867--3874. IEEE, 2023.

\bibitem[Lin et~al.(2024{\natexlab{a}})Lin, Yin, Qi, Abbeel, and
  Malik]{lin2024twisting}
Toru Lin, Zhao-Heng Yin, Haozhi Qi, Pieter Abbeel, and Jitendra Malik.
\newblock Twisting lids off with two hands.
\newblock \emph{arXiv preprint arXiv:2403.02338}, 2024{\natexlab{a}}.

\bibitem[Lin et~al.(2024{\natexlab{b}})Lin, Zhang, Li, Qi, Yi, Levine, and
  Malik]{lin2024learning}
Toru Lin, Yu~Zhang, Qiyang Li, Haozhi Qi, Brent Yi, Sergey Levine, and Jitendra
  Malik.
\newblock Learning visuotactile skills with two multifingered hands.
\newblock \emph{arXiv preprint arXiv:2404.16823}, 2024{\natexlab{b}}.

\bibitem[Liu et~al.(2024{\natexlab{a}})Liu, He, Seita, and
  Sukhatme]{liu2024voxact}
I-Chun~Arthur Liu, Sicheng He, Daniel Seita, and Gaurav~S Sukhatme.
\newblock Voxact-b: Voxel-based acting and stabilizing policy for bimanual
  manipulation.
\newblock In \emph{8th Annual Conference on Robot Learning},
  2024{\natexlab{a}}.

\bibitem[Liu et~al.(2022)Liu, Chen, Dong, Wang, Calinon, Li, and
  Chen]{liu2022robot}
Junjia Liu, Yiting Chen, Zhipeng Dong, Shixiong Wang, Sylvain Calinon, Miao Li,
  and Fei Chen.
\newblock Robot cooking with stir-fry: Bimanual non-prehensile manipulation of
  semi-fluid objects.
\newblock \emph{IEEE Robotics and Automation Letters}, 7\penalty0 (2):\penalty0
  5159--5166, 2022.

\bibitem[Liu et~al.(2024{\natexlab{b}})Liu, Wu, Li, Tan, Chen, Wang, Xu, Su,
  and Zhu]{liu2024rdt}
Songming Liu, Lingxuan Wu, Bangguo Li, Hengkai Tan, Huayu Chen, Zhengyi Wang,
  Ke~Xu, Hang Su, and Jun Zhu.
\newblock Rdt-1b: a diffusion foundation model for bimanual manipulation.
\newblock \emph{arXiv preprint arXiv:2410.07864}, 2024{\natexlab{b}}.

\bibitem[Liu et~al.(2024{\natexlab{c}})Liu, Yang, Si, Liu, Li, Zhang, Liu, and
  Yi]{liu2024taco}
Yun Liu, Haolin Yang, Xu~Si, Ling Liu, Zipeng Li, Yuxiang Zhang, Yebin Liu, and
  Li~Yi.
\newblock Taco: Benchmarking generalizable bimanual tool-action-object
  understanding.
\newblock In \emph{Proceedings of the IEEE/CVF Conference on Computer Vision
  and Pattern Recognition}, pages 21740--21751, 2024{\natexlab{c}}.

\bibitem[Ma et~al.(2024)Ma, Patidar, Haughton, and James]{ma2024hierarchical}
Xiao Ma, Sumit Patidar, Iain Haughton, and Stephen James.
\newblock Hierarchical diffusion policy for kinematics-aware multi-task robotic
  manipulation.
\newblock In \emph{Proceedings of the IEEE/CVF Conference on Computer Vision
  and Pattern Recognition}, pages 18081--18090, 2024.

\bibitem[Maitin-Shepard et~al.(2010)Maitin-Shepard, Cusumano-Towner, Lei, and
  Abbeel]{maitin2010cloth}
Jeremy Maitin-Shepard, Marco Cusumano-Towner, Jinna Lei, and Pieter Abbeel.
\newblock Cloth grasp point detection based on multiple-view geometric cues
  with application to robotic towel folding.
\newblock In \emph{2010 IEEE International Conference on Robotics and
  Automation}, pages 2308--2315. IEEE, 2010.

\bibitem[Mandlekar et~al.(2020)Mandlekar, Xu, Mart{\'\i}n-Mart{\'\i}n,
  Savarese, and Fei-Fei]{mandlekar2020learning}
Ajay Mandlekar, Danfei Xu, Roberto Mart{\'\i}n-Mart{\'\i}n, Silvio Savarese,
  and Li~Fei-Fei.
\newblock Learning to generalize across long-horizon tasks from human
  demonstrations.
\newblock In \emph{Proceedings of Robotics: Science and Systems (RSS)}, 2020.

\bibitem[Mandlekar et~al.(2022)Mandlekar, Xu, Wong, Nasiriany, Wang, Kulkarni,
  Fei-Fei, Savarese, Zhu, and Mart{\'\i}n-Mart{\'\i}n]{mandlekar2022matters}
Ajay Mandlekar, Danfei Xu, Josiah Wong, Soroush Nasiriany, Chen Wang, Rohun
  Kulkarni, Li~Fei-Fei, Silvio Savarese, Yuke Zhu, and Roberto
  Mart{\'\i}n-Mart{\'\i}n.
\newblock What matters in learning from offline human demonstrations for robot
  manipulation.
\newblock In \emph{Conference on Robot Learning}, pages 1678--1690. PMLR, 2022.

\bibitem[Mees et~al.(2022)Mees, Hermann, Rosete-Beas, and
  Burgard]{mees2022calvin}
Oier Mees, Lukas Hermann, Erick Rosete-Beas, and Wolfram Burgard.
\newblock Calvin: A benchmark for language-conditioned policy learning for
  long-horizon robot manipulation tasks.
\newblock \emph{IEEE Robotics and Automation Letters}, 7\penalty0 (3):\penalty0
  7327--7334, 2022.

\bibitem[Miller and Allen(2004)]{miller2004graspit}
Andrew~T Miller and Peter~K Allen.
\newblock Graspit! a versatile simulator for robotic grasping.
\newblock \emph{IEEE Robotics \& Automation Magazine}, 11\penalty0
  (4):\penalty0 110--122, 2004.

\bibitem[Mirrazavi~Salehian et~al.(2016)Mirrazavi~Salehian, Figueroa~Fernandez,
  and Billard]{mirrazavi2016coordinated}
Seyed~Sina Mirrazavi~Salehian, Nadia~Barbara Figueroa~Fernandez, and Aude
  Billard.
\newblock Coordinated multi-arm motion planning: Reaching for moving objects in
  the face of uncertainty.
\newblock In \emph{Proceedings of Robotics: Science and Systems (RSS)}, 2016.

\bibitem[Mu et~al.(2024)Mu, Chen, Peng, Chen, Gao, Zou, Lin, Xie, and
  Luo]{mu2024robotwin}
Yao Mu, Tianxing Chen, Shijia Peng, Zanxin Chen, Zeyu Gao, Yude Zou, Lunkai
  Lin, Zhiqiang Xie, and Ping Luo.
\newblock Robotwin: Dual-arm robot benchmark with generative digital twins
  (early version).
\newblock \emph{arXiv preprint arXiv:2409.02920}, 2024.

\bibitem[Nasiriany et~al.(2024)Nasiriany, Kirmani, Ding, Smith, Zhu, Driess,
  Sadigh, and Xiao]{nasiriany2024rt}
Soroush Nasiriany, Sean Kirmani, Tianli Ding, Laura Smith, Yuke Zhu, Danny
  Driess, Dorsa Sadigh, and Ted Xiao.
\newblock Rt-affordance: Affordances are versatile intermediate representations
  for robot manipulation.
\newblock \emph{arXiv preprint arXiv:2411.02704}, 2024.

\bibitem[Nichol and Dhariwal(2021)]{nichol2021improved}
Alexander~Quinn Nichol and Prafulla Dhariwal.
\newblock Improved denoising diffusion probabilistic models.
\newblock In \emph{International Conference on Machine Learning}, pages
  8162--8171. PMLR, 2021.

\bibitem[O’Neill et~al.(2024)O’Neill, Rehman, Maddukuri, Gupta, Padalkar,
  Lee, Pooley, Gupta, Mandlekar, Jain, et~al.]{o2024open}
Abby O’Neill, Abdul Rehman, Abhiram Maddukuri, Abhishek Gupta, Abhishek
  Padalkar, Abraham Lee, Acorn Pooley, Agrim Gupta, Ajay Mandlekar, Ajinkya
  Jain, et~al.
\newblock Open x-embodiment: Robotic learning datasets and rt-x models: Open
  x-embodiment collaboration 0.
\newblock In \emph{2024 IEEE International Conference on Robotics and
  Automation (ICRA)}, pages 6892--6903. IEEE, 2024.

\bibitem[Papagiannis et~al.(2024)Papagiannis, Di~Palo, Vitiello, and
  Johns]{papagiannis2024rx}
Georgios Papagiannis, Norman Di~Palo, Pietro Vitiello, and Edward Johns.
\newblock R+x: Retrieval and execution from everyday human videos.
\newblock \emph{arXiv preprint arXiv:2407.12957}, 2024.

\bibitem[Pavlakos et~al.(2024)Pavlakos, Shan, Radosavovic, Kanazawa, Fouhey,
  and Malik]{pavlakos2024reconstructing}
Georgios Pavlakos, Dandan Shan, Ilija Radosavovic, Angjoo Kanazawa, David
  Fouhey, and Jitendra Malik.
\newblock Reconstructing hands in 3d with transformers.
\newblock In \emph{Proceedings of the IEEE/CVF Conference on Computer Vision
  and Pattern Recognition}, pages 9826--9836, 2024.

\bibitem[Peer et~al.(2007)Peer, Komoguchi, and Buss]{peer2007towards}
Angelika Peer, Yuta Komoguchi, and Martin Buss.
\newblock Towards a mobile haptic interface for bimanual manipulations.
\newblock In \emph{2007 IEEE/RSJ International Conference on Intelligent Robots
  and Systems}, pages 384--391. IEEE, 2007.

\bibitem[Peng et~al.(2024)Peng, Lv, Zeng, Chen, Zhao, Sun, Lu, and
  Shao]{peng2024tiebot}
Weikun Peng, Jun Lv, Yuwei Zeng, Haonan Chen, Siheng Zhao, Jichen Sun, Cewu Lu,
  and Lin Shao.
\newblock Tiebot: Learning to knot a tie from visual demonstration through a
  real-to-sim-to-real approach.
\newblock In \emph{8th Annual Conference on Robot Learning}, 2024.

\bibitem[Perez et~al.(2018)Perez, Strub, De~Vries, Dumoulin, and
  Courville]{perez2018film}
Ethan Perez, Florian Strub, Harm De~Vries, Vincent Dumoulin, and Aaron
  Courville.
\newblock Film: Visual reasoning with a general conditioning layer.
\newblock In \emph{Proceedings of the AAAI conference on artificial
  intelligence}, volume~32, 2018.

\bibitem[Potamias et~al.(2024)Potamias, Zhang, Deng, and
  Zafeiriou]{potamias2024wilor}
Rolandos~Alexandros Potamias, Jinglei Zhang, Jiankang Deng, and Stefanos
  Zafeiriou.
\newblock Wilor: End-to-end 3d hand localization and reconstruction
  in-the-wild.
\newblock \emph{arXiv preprint arXiv:2409.12259}, 2024.

\bibitem[Qi et~al.(2017)Qi, Yi, Su, and Guibas]{qi2017pointnet}
Charles~Ruizhongtai Qi, Li~Yi, Hao Su, and Leonidas~J Guibas.
\newblock Pointnet++: Deep hierarchical feature learning on point sets in a
  metric space.
\newblock \emph{Advances in Neural Information Processing Systems}, 30, 2017.

\bibitem[Ravi et~al.(2024)Ravi, Gabeur, Hu, Hu, Ryali, Ma, Khedr, R{\"a}dle,
  Rolland, Gustafson, et~al.]{ravi2024sam}
Nikhila Ravi, Valentin Gabeur, Yuan-Ting Hu, Ronghang Hu, Chaitanya Ryali,
  Tengyu Ma, Haitham Khedr, Roman R{\"a}dle, Chloe Rolland, Laura Gustafson,
  et~al.
\newblock Sam 2: Segment anything in images and videos.
\newblock \emph{arXiv preprint arXiv:2408.00714}, 2024.

\bibitem[Romero et~al.(2017)Romero, Tzionas, and Black]{romero2017embodied}
Javier Romero, Dimitris Tzionas, and Michael~J Black.
\newblock Embodied hands: Modeling and capturing hands and bodies together.
\newblock \emph{ACM Transactions on Graphics}, 36\penalty0 (6), 2017.

\bibitem[Ronneberger et~al.(2015)Ronneberger, Fischer, and
  Brox]{ronneberger2015u}
Olaf Ronneberger, Philipp Fischer, and Thomas Brox.
\newblock U-net: Convolutional networks for biomedical image segmentation.
\newblock In \emph{Medical Image Computing and Computer-Assisted
  Intervention--MICCAI}, pages 234--241. Springer, 2015.

\bibitem[Ryu et~al.(2024)Ryu, Kim, An, Chang, Seo, Kim, Kim, Hwang, Choi, and
  Horowitz]{ryu2024diffusion}
Hyunwoo Ryu, Jiwoo Kim, Hyunseok An, Junwoo Chang, Joohwan Seo, Taehan Kim,
  Yubin Kim, Chaewon Hwang, Jongeun Choi, and Roberto Horowitz.
\newblock Diffusion-edfs: Bi-equivariant denoising generative modeling on se
  (3) for visual robotic manipulation.
\newblock In \emph{Proceedings of the IEEE/CVF Conference on Computer Vision
  and Pattern Recognition}, pages 18007--18018, 2024.

\bibitem[Salhotra et~al.(2023)Salhotra, Liu, and
  Sukhatme]{salhotra2023learning}
Gautam Salhotra, I-Chun~Arthur Liu, and Gaurav~S Sukhatme.
\newblock Learning robot manipulation from cross-morphology demonstration.
\newblock In \emph{Conference on Robot Learning}, pages 2257--2277. PMLR, 2023.

\bibitem[Shan et~al.(2020)Shan, Geng, Shu, and Fouhey]{shan2020understanding}
Dandan Shan, Jiaqi Geng, Michelle Shu, and David~F Fouhey.
\newblock Understanding human hands in contact at internet scale.
\newblock In \emph{Proceedings of the IEEE/CVF Conference on Computer Vision
  and Pattern Recognition}, pages 9869--9878, 2020.

\bibitem[Shaw et~al.(2024)Shaw, Li, Yang, Srirama, Liu, Xiong, Mendonca, and
  Pathak]{shaw2024bimanual}
Kenneth Shaw, Yulong Li, Jiahui Yang, Mohan~Kumar Srirama, Ray Liu, Haoyu
  Xiong, Russell Mendonca, and Deepak Pathak.
\newblock Bimanual dexterity for complex tasks.
\newblock In \emph{8th Annual Conference on Robot Learning}, 2024.

\bibitem[Shridhar et~al.(2023)Shridhar, Manuelli, and
  Fox]{shridhar2023perceiver}
Mohit Shridhar, Lucas Manuelli, and Dieter Fox.
\newblock Perceiver-actor: A multi-task transformer for robotic manipulation.
\newblock In \emph{Conference on Robot Learning}, pages 785--799. PMLR, 2023.

\bibitem[Sirintuna et~al.(2023)Sirintuna, Ozdamar, and
  Ajoudani]{sirintuna2023carrying}
Doganay Sirintuna, Idil Ozdamar, and Arash Ajoudani.
\newblock Carrying the uncarriable: a deformation-agnostic and
  human-cooperative framework for unwieldy objects using multiple robots.
\newblock In \emph{2023 IEEE International Conference on Robotics and
  Automation (ICRA)}, pages 7497--7503. IEEE, 2023.

\bibitem[Smith et~al.(2012)Smith, Karayiannidis, Nalpantidis, Gratal, Qi,
  Dimarogonas, and Kragic]{smith2012dual}
Christian Smith, Yiannis Karayiannidis, Lazaros Nalpantidis, Xavi Gratal, Peng
  Qi, Dimos~V Dimarogonas, and Danica Kragic.
\newblock Dual arm manipulation—a survey.
\newblock \emph{Robotics and Autonomous systems}, 60\penalty0 (10):\penalty0
  1340--1353, 2012.

\bibitem[Song et~al.(2021)Song, Meng, and Ermon]{song2021denoising}
Jiaming Song, Chenlin Meng, and Stefano Ermon.
\newblock Denoising diffusion implicit models.
\newblock In \emph{International Conference on Learning Representations}, 2021.

\bibitem[Team et~al.(2024)Team, Ghosh, Walke, Pertsch, Black, Mees, Dasari,
  Hejna, Kreiman, Xu, et~al.]{team2024octo}
Octo~Model Team, Dibya Ghosh, Homer Walke, Karl Pertsch, Kevin Black, Oier
  Mees, Sudeep Dasari, Joey Hejna, Tobias Kreiman, Charles Xu, et~al.
\newblock Octo: An open-source generalist robot policy.
\newblock \emph{arXiv preprint arXiv:2405.12213}, 2024.

\bibitem[Wang et~al.(2023)Wang, Fan, Sun, Zhang, Fei-Fei, Xu, Zhu, and
  Anandkumar]{wang2023mimicplay}
Chen Wang, Linxi Fan, Jiankai Sun, Ruohan Zhang, Li~Fei-Fei, Danfei Xu, Yuke
  Zhu, and Anima Anandkumar.
\newblock Mimicplay: Long-horizon imitation learning by watching human play.
\newblock In \emph{Conference on Robot Learning}, pages 201--221. PMLR, 2023.

\bibitem[Wang et~al.(2024)Wang, Shi, Wang, Zhang, Fei-Fei, and
  Liu]{wang2024dexcap}
Chen Wang, Haochen Shi, Weizhuo Wang, Ruohan Zhang, Li~Fei-Fei, and C~Karen
  Liu.
\newblock Dexcap: Scalable and portable mocap data collection system for
  dexterous manipulation.
\newblock In \emph{Proceedings of Robotics: Science and Systems (RSS)}, 2024.

\bibitem[Wen et~al.(2023)Wen, Lin, So, Chen, Dou, Gao, and Abbeel]{wen2023any}
Chuan Wen, Xingyu Lin, John So, Kai Chen, Qi~Dou, Yang Gao, and Pieter Abbeel.
\newblock Any-point trajectory modeling for policy learning.
\newblock \emph{arXiv preprint arXiv:2401.00025}, 2023.

\bibitem[Weng et~al.(2022)Weng, Bajracharya, Wang, Agrawal, and
  Held]{weng2022fabricflownet}
Thomas Weng, Sujay~Man Bajracharya, Yufei Wang, Khush Agrawal, and David Held.
\newblock Fabricflownet: Bimanual cloth manipulation with a flow-based policy.
\newblock In \emph{Conference on Robot Learning}, pages 192--202. PMLR, 2022.

\bibitem[Xian et~al.(2023)Xian, Gkanatsios, Gervet, Ke, and
  Fragkiadaki]{xian2023chaineddiffuser}
Zhou Xian, Nikolaos Gkanatsios, Theophile Gervet, Tsung-Wei Ke, and Katerina
  Fragkiadaki.
\newblock Chaineddiffuser: Unifying trajectory diffusion and keypose prediction
  for robotic manipulation.
\newblock In \emph{7th Annual Conference on Robot Learning}, 2023.

\bibitem[Xiao et~al.(2024)Xiao, Wu, Xu, Dai, Hu, Lu, Zeng, Liu, and
  Yuan]{xiao2024florence}
Bin Xiao, Haiping Wu, Weijian Xu, Xiyang Dai, Houdong Hu, Yumao Lu, Michael
  Zeng, Ce~Liu, and Lu~Yuan.
\newblock Florence-2: Advancing a unified representation for a variety of
  vision tasks.
\newblock In \emph{Proceedings of the IEEE/CVF Conference on Computer Vision
  and Pattern Recognition}, pages 4818--4829, 2024.

\bibitem[Xie et~al.(2020)Xie, Chowdhury, De~Paolis~Kaluza, Zhao, Wong, and
  Yu]{xie2020deep}
Fan Xie, Alexander Chowdhury, M~De~Paolis~Kaluza, Linfeng Zhao, Lawson Wong,
  and Rose Yu.
\newblock Deep imitation learning for bimanual robotic manipulation.
\newblock \emph{Advances in Neural Information Processing Systems},
  33:\penalty0 2327--2337, 2020.

\bibitem[Xu et~al.(2023)Xu, Wang, Ding, and Yang]{xu2023iterative}
Gangwei Xu, Xianqi Wang, Xiaohuan Ding, and Xin Yang.
\newblock Iterative geometry encoding volume for stereo matching.
\newblock In \emph{Proceedings of the IEEE/CVF Conference on Computer Vision
  and Pattern Recognition}, pages 21919--21928, 2023.

\bibitem[Xu et~al.(2024)Xu, Wang, Zhang, Cheng, Liao, and Yang]{xu2024igev}
Gangwei Xu, Xianqi Wang, Zhaoxing Zhang, Junda Cheng, Chunyuan Liao, and Xin
  Yang.
\newblock Igev++: Iterative multi-range geometry encoding volumes for stereo
  matching.
\newblock \emph{arXiv preprint arXiv:2409.00638}, 2024.

\bibitem[Yan et~al.(2024)Yan, Stouraitis, Moura, Xu, Gienger, and
  Vijayakumar]{yan2024impact}
Lei Yan, Theodoros Stouraitis, Jo{\~a}o Moura, Wenfu Xu, Michael Gienger, and
  Sethu Vijayakumar.
\newblock Impact-aware bimanual catching of large-momentum objects.
\newblock \emph{IEEE Transactions on Robotics}, 2024.

\bibitem[Yang et~al.(2024{\natexlab{a}})Yang, Cao, Deng, Antonova, Song, and
  Bohg]{yange2024quibot}
Jingyun Yang, Ziang Cao, Congyue Deng, Rika Antonova, Shuran Song, and
  Jeannette Bohg.
\newblock Equibot: Sim (3)-equivariant diffusion policy for generalizable and
  data efficient learning.
\newblock In \emph{8th Annual Conference on Robot Learning},
  2024{\natexlab{a}}.

\bibitem[Yang et~al.(2024{\natexlab{b}})Yang, Deng, Wu, Antonova, Guibas, and
  Bohg]{yang2024equivact}
Jingyun Yang, Congyue Deng, Jimmy Wu, Rika Antonova, Leonidas Guibas, and
  Jeannette Bohg.
\newblock Equivact: Sim (3)-equivariant visuomotor policies beyond rigid object
  manipulation.
\newblock In \emph{2024 IEEE International Conference on Robotics and
  Automation (ICRA)}, pages 9249--9255. IEEE, 2024{\natexlab{b}}.

\bibitem[Yang et~al.(2024{\natexlab{c}})Yang, Han, and
  Ravichandar]{yang2024asymdex}
Zhaodong Yang, Yunhai Han, and Harish Ravichandar.
\newblock Asymdex: Leveraging asymmetry and relative motion in learning
  bimanual dexterity.
\newblock \emph{arXiv preprint arXiv:2411.13020}, 2024{\natexlab{c}}.

\bibitem[Ze et~al.(2024)Ze, Zhang, Zhang, Hu, Wang, and Xu]{ze2024dp3}
Yanjie Ze, Gu~Zhang, Kangning Zhang, Chenyuan Hu, Muhan Wang, and Huazhe Xu.
\newblock 3d diffusion policy: Generalizable visuomotor policy learning via
  simple 3d representations.
\newblock In \emph{Proceedings of Robotics: Science and Systems (RSS)}, 2024.

\bibitem[Zeng et~al.(2024)Zeng, Bu, Wang, Xia, Chen, Dong, Song, Wang, Hu, Luo,
  et~al.]{zeng2024learning}
Jia Zeng, Qingwen Bu, Bangjun Wang, Wenke Xia, Li~Chen, Hao Dong, Haoming Song,
  Dong Wang, Di~Hu, Ping Luo, et~al.
\newblock Learning manipulation by predicting interaction.
\newblock In \emph{Proceedings of Robotics: Science and Systems (RSS)}, 2024.

\bibitem[Zhan et~al.(2024)Zhan, Yang, Zhao, Mao, Xu, Lin, Li, and
  Lu]{zhan2024oakink2}
Xinyu Zhan, Lixin Yang, Yifei Zhao, Kangrui Mao, Hanlin Xu, Zenan Lin, Kailin
  Li, and Cewu Lu.
\newblock Oakink2: A dataset of bimanual hands-object manipulation in complex
  task completion.
\newblock In \emph{Proceedings of the IEEE/CVF Conference on Computer Vision
  and Pattern Recognition}, pages 445--456, 2024.

\bibitem[Zhang and Gienger(2024)]{zhang2024affordance}
Fan Zhang and Michael Gienger.
\newblock Affordance-based robot manipulation with flow matching.
\newblock \emph{arXiv preprint arXiv:2409.01083}, 2024.

\bibitem[Zhang et~al.(2024)Zhang, Li, Li, Zeng, Zhao, Sun, Chen, Wei, Zhan, Li,
  et~al.]{zhang2024empowering}
Tianle Zhang, Dongjiang Li, Yihang Li, Zecui Zeng, Lin Zhao, Lei Sun, Yue Chen,
  Xuelong Wei, Yibing Zhan, Lusong Li, et~al.
\newblock Empowering embodied manipulation: A bimanual-mobile robot
  manipulation dataset for household tasks.
\newblock \emph{arXiv preprint arXiv:2405.18860}, 2024.

\bibitem[Zhang and Boularias(2024)]{zhang2024one}
Xinyu Zhang and Abdeslam Boularias.
\newblock One-shot imitation learning with invariance matching for robotic
  manipulation.
\newblock In \emph{Proceedings of Robotics: Science and Systems (RSS)}, 2024.

\bibitem[Zhao et~al.(2023{\natexlab{a}})Zhao, Kumar, Levine, and
  Finn]{zhao2023learning}
Tony~Z Zhao, Vikash Kumar, Sergey Levine, and Chelsea Finn.
\newblock Learning fine-grained bimanual manipulation with low-cost hardware.
\newblock In \emph{Proceedings of Robotics: Science and Systems (RSS)},
  2023{\natexlab{a}}.

\bibitem[Zhao et~al.(2024)Zhao, Tompson, Driess, Florence, Ghasemipour, Finn,
  and Wahid]{zhao2024aloha}
Tony~Z Zhao, Jonathan Tompson, Danny Driess, Pete Florence, Seyed Kamyar~Seyed
  Ghasemipour, Chelsea Finn, and Ayzaan Wahid.
\newblock Aloha unleashed: A simple recipe for robot dexterity.
\newblock In \emph{8th Annual Conference on Robot Learning}, 2024.

\bibitem[Zhao et~al.(2023{\natexlab{b}})Zhao, Wu, Chen, Zhang, Fan, Mo, and
  Dong]{zhaodual2023afford}
Yan Zhao, Ruihai Wu, Zhehuan Chen, Yourong Zhang, Qingnan Fan, Kaichun Mo, and
  Hao Dong.
\newblock Dualafford: Learning collaborative visual affordance for dual-gripper
  manipulation.
\newblock In \emph{The Eleventh International Conference on Learning
  Representations}, 2023{\natexlab{b}}.

\bibitem[Zhou et~al.(2024)Zhou, Yuan, Fu, and Lu]{zhou2024learning}
Bohan Zhou, Haoqi Yuan, Yuhui Fu, and Zongqing Lu.
\newblock Learning diverse bimanual dexterous manipulation skills from human
  demonstrations.
\newblock \emph{arXiv preprint arXiv:2410.02477}, 2024.

\bibitem[Zhu et~al.(2024)Zhu, Gienger, Franzese, and Kober]{zhu2024you}
Jihong Zhu, Michael Gienger, Giovanni Franzese, and Jens Kober.
\newblock Do you need a hand?--a bimanual robotic dressing assistance scheme.
\newblock \emph{IEEE Transactions on Robotics}, 40:\penalty0 1906--1919, 2024.

\end{thebibliography}


\clearpage
\appendix


\subsection{Implementation Details of Our BiDP}



In this section, we describe in detail the architecture and implementation of our proposed method BiDP.

\subsubsection{Spaces of observation and action}
We adopt a 13-dimensional proprioception vector and a 7-dimensional action space for each robot arm, respectively. The proprioception data for each arm consists of the following information: a 3-dimensional end-effector position, a 6-dimensional vector denoting end-effector orientation (represented by two columns of the end-effector rotation matrix), a 3-dimensional vector indicating the direction of gravity, and a scalar that represents the degree to which the gripper is opened. The action space for each arm consists of the following information: a 3-dimensional vector for the end-effector position offset, a 3-dimensional vector for the end-effector angular velocity in axis-angle format, and a scalar denoting the gripper action.

For all our bimanual tasks, the observation horizon is set to 1, so we only use the initial state observation of the left arm as one of the network inputs. And the initial state of the right arm is always fixed in each task. For the number of the action steps, which is also the length of the predicted horizon, we simplify it and set the prediction length of the three strictly asynchronous tasks to the number of keyframes $K$, and the prediction length of the two synchronous tasks to $2K$, which is not reducible. This is slightly different from the setup used in the mainstream methods ACT \cite{zhao2023learning}, Diffusion Policy \cite{chi2023diffusion} and EquiBot \cite{yange2024quibot}, where the action horizon is always smaller than the prediction horizon with redundant steps.

\subsubsection{Network architecture}
In all tasks, we use a SIM(3)-equivariant PointNet++ \cite{yang2024equivact, yange2024quibot} with 4 layers and hidden dimensionality 128 as the feature encoder. For the noise prediction network, we inherits hyperparameters from the original Diffusion Policy \cite{chi2023diffusion}. Specifically, to optimize for inference speed  in all experiments, we use the DDIM scheduler \cite{song2021denoising} with 8 denoising steps, instead of the DDPM scheduler \cite{ho2020denoising} which performs up to 100 denoising steps.

\subsubsection{Sampling of point cloud}
As we all known, setting the number of points to sample in the point cloud observation is a key hyperparameter to consider when designing an architecture that takes point cloud inputs. In our experiments, we found out that using 1024 points is sufficient for all tasks. In particular, we have tried increasing the number of point clouds to 2048 or more, but the evaluation improvement in each task is minimal, and this will also cause the storage occupied by the training observation data to be too large and the training time cost to increase. Therefore, reducing the number of points to 1024 can make training faster without hurting performance. And all our policy models can be trained on a GeForce RTX 3090 Ti with 24 GB of memory.

\begin{figure}[]
	\includegraphics[width=0.495\columnwidth]{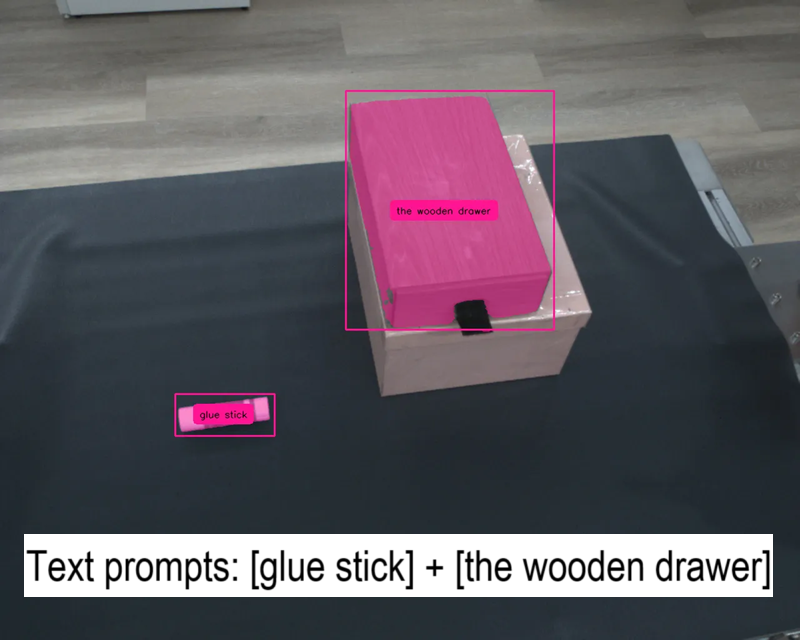}
 	\includegraphics[width=0.495\columnwidth]{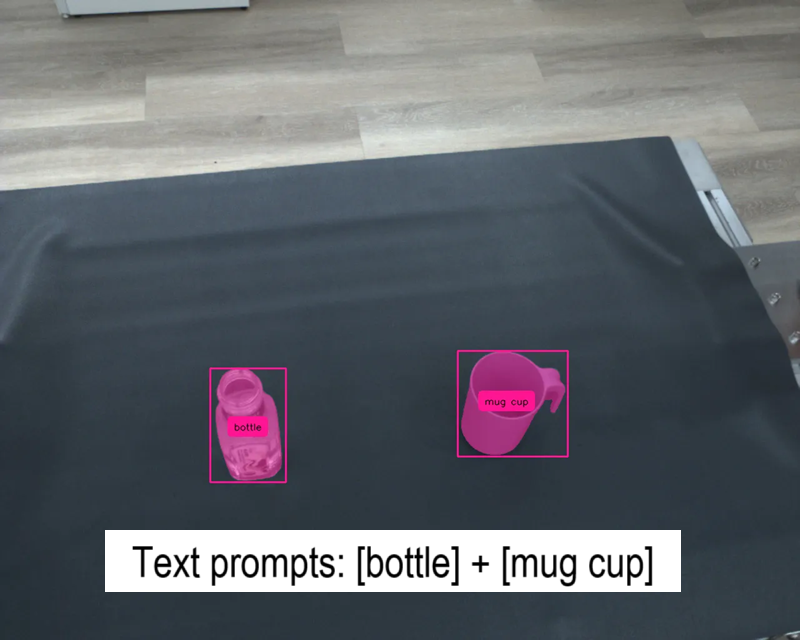}\\
	\vspace{-10pt}\\
	\includegraphics[width=0.495\columnwidth]{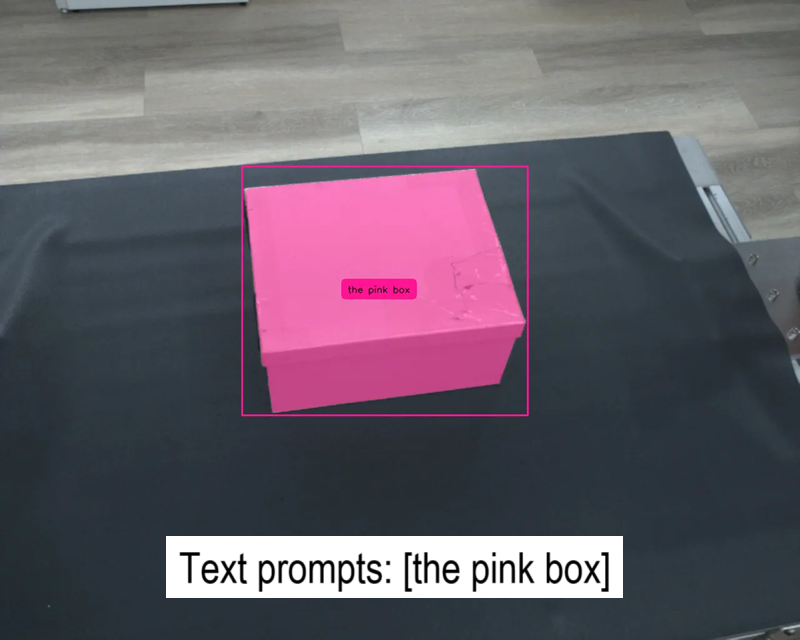}
 	\includegraphics[width=0.495\columnwidth]{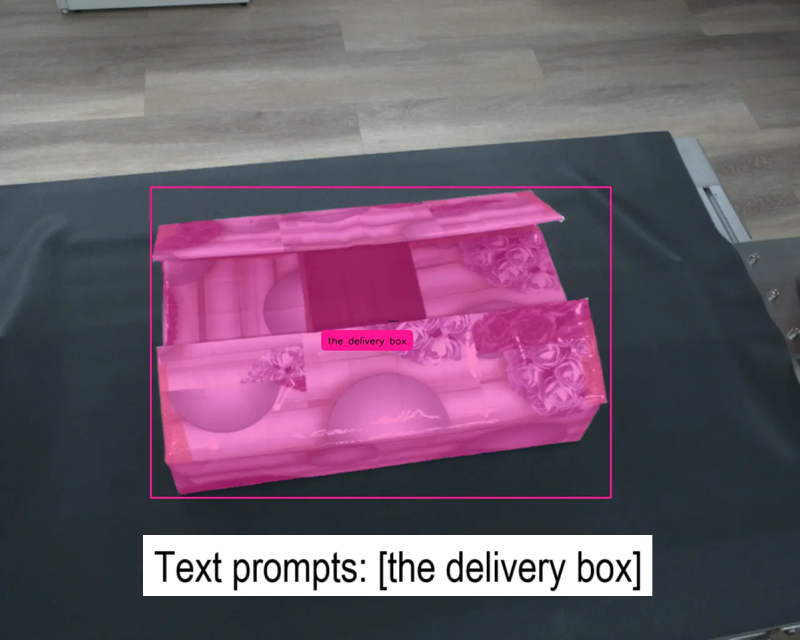}
	\caption{Examples of using vision foundation models (VFMs) to detect and segment manipulated objects.}
	\label{suppVFMs}
	\vspace{-15pt}
\end{figure}

\subsubsection{Training and evaluation}
When using fully expanded training demonstrations (including $\times100$ and $\times500$), we train all methods of the first two tasks and the last three tasks for 500 and 1,000 epochs, respectively. The batch-size is set to 64. Otherwise, when using these under-expanded training data (including $\times25$ and $\times5$ and not expanded), we train all methods of the first two tasks and the last three tasks for 2,000 and 4,000 epochs, respectively. For all experiments, we only evaluate the last one checkpoint saved at the end of training. For every evaluation in the real world, we run the policy in a randomly initialized placement of objects for dozens of episodes (please refer the metrics part in the main content for more details), and record the mean average length and success rate achieved by the policy.

In addition, we have explained and demonstrated the importance and advantages of object-centric point cloud input in our main paper. At inference time, we also need to preprocess the binocular RGB observations to obtain the point cloud of manipulated objects. This core design relies on the still rapidly developing capabilities of vision foundation models (VFMs). Here we leverage the state-of-the-art open vocabulary detection method Florence-2 \cite{xiao2024florence} and segmentation method SAM2 \cite{ravi2024sam} to automatically extract object masks and then filter out corresponding point clouds. Examples are shown in Fig.~\ref{suppVFMs}. Despite this, occasionally we may fail to segment desired objects accurately, and in these special cases we will manually correct the masks. These cases are not counted as failed evaluation trails due to not involving significant elements of bimanual robot manipulation. Because we believe that the next generation of VFMs can alleviate these problems, or we can directly address them through domain adaptation, test-time adaptation, or adjusting input prompts.

\begin{figure*}[]
	\includegraphics[width=0.495\textwidth]{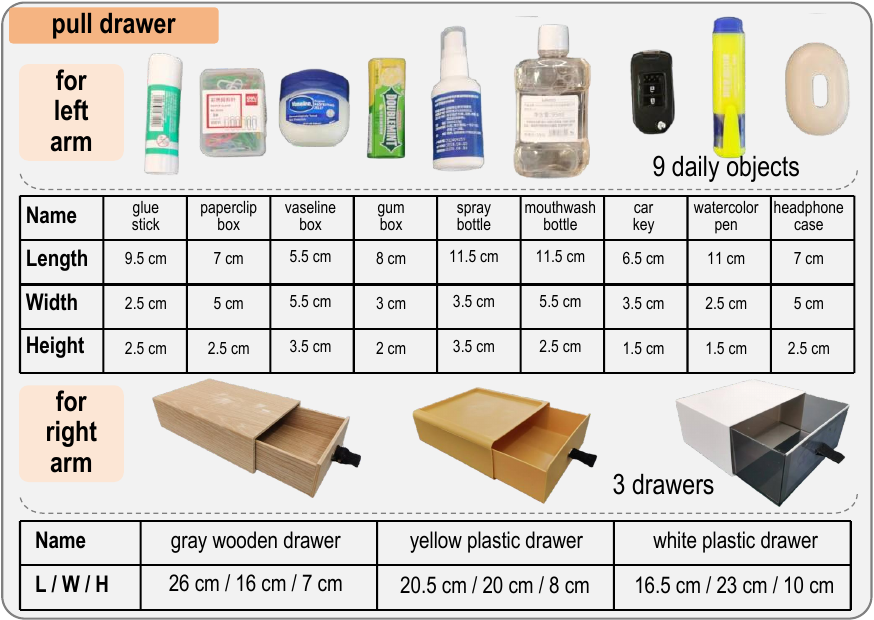}
 	\includegraphics[width=0.495\textwidth]{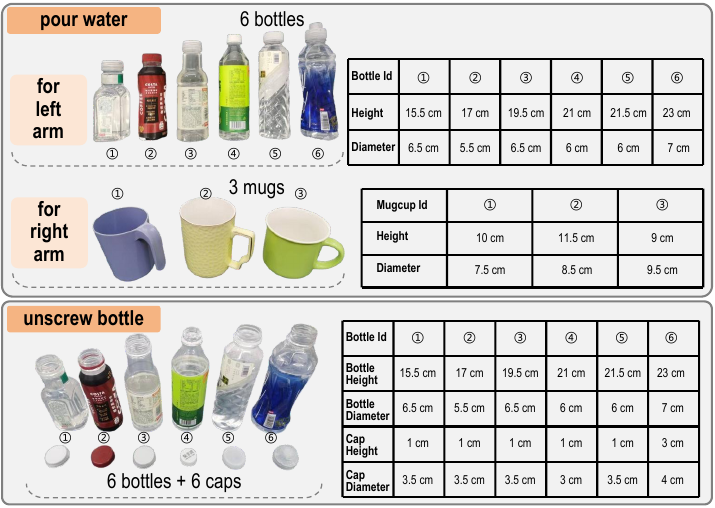}\\
	\vspace{-10pt}\\
 	\includegraphics[width=0.505\textwidth]{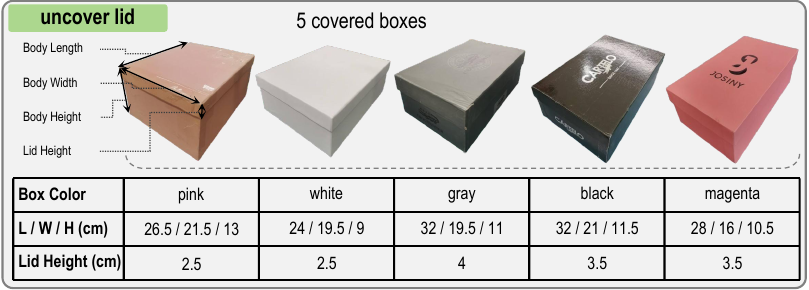}
 	\includegraphics[width=0.485\textwidth]{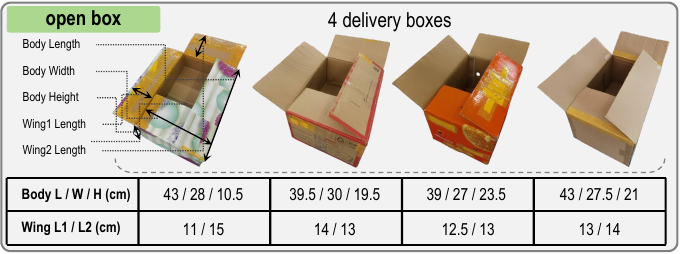}
	\caption{We collected a variety of manipulated objects in instance-level for each of five bimanual tasks to improve and verify the generalizability of trained policies. All of these objects are from everyday life, not intentionally customized. We also collect the detailed size information in centimeter-level for all related objects. Best to view after zooming in.}
	\label{suppAssets}
	\vspace{-10pt}
\end{figure*}

\begin{figure*}[]
	\centering
 	\includegraphics[width=0.9\textwidth]{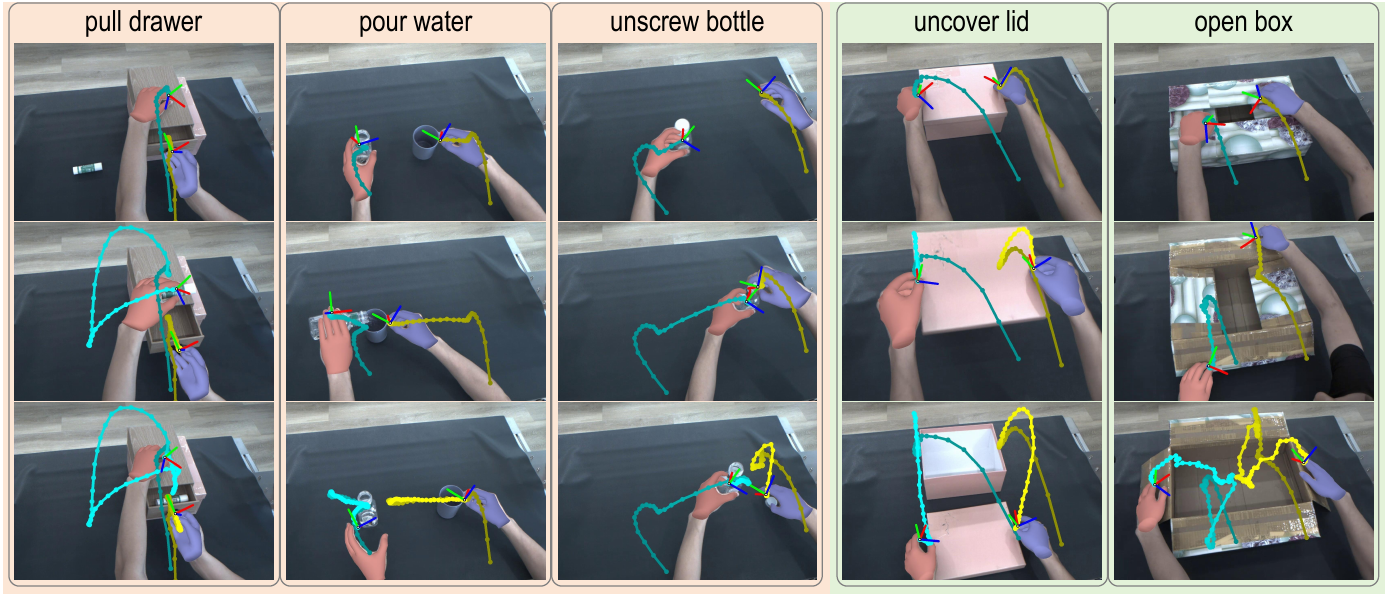}
	\caption{Visualization of extracted hand trajectories for five long-horizon bimanual tasks. Best to view after zooming in.}
	\label{tasksTrajs}
	\vspace{-10pt}
\end{figure*}

\begin{figure*}[]
	\includegraphics[width=0.333\textwidth]{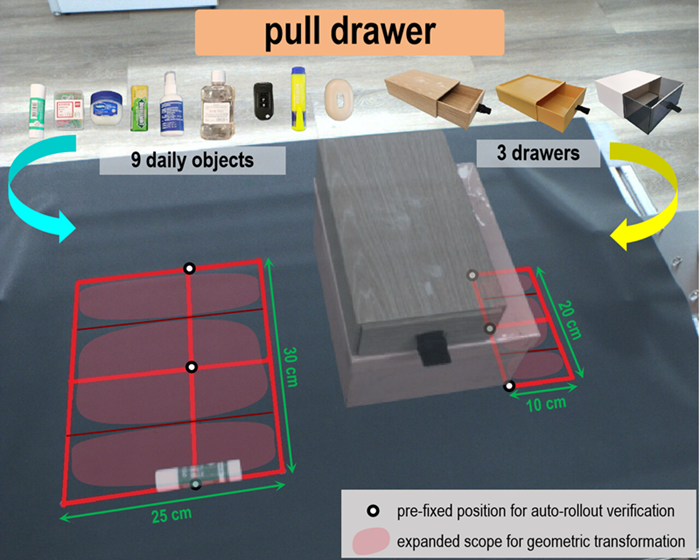}
 	\includegraphics[width=0.333\textwidth]{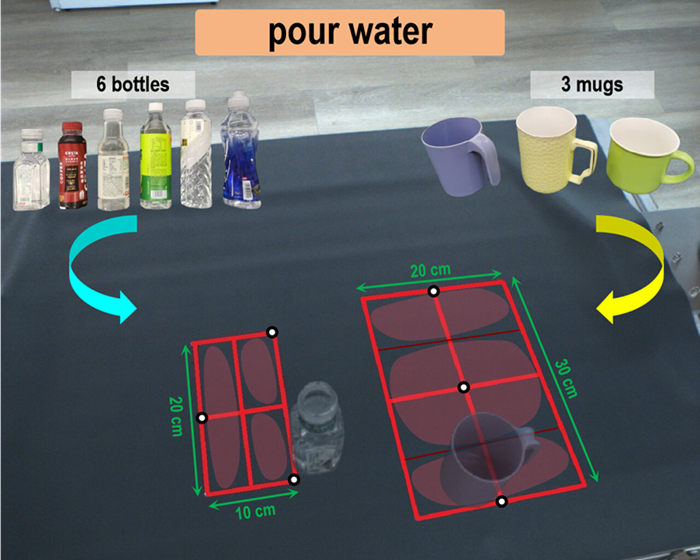}
 	\includegraphics[width=0.333\textwidth]{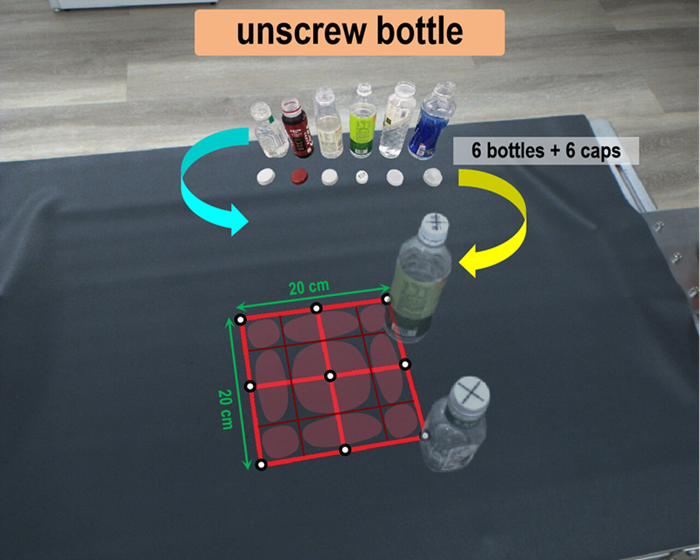}\\
	\vspace{-10pt}\\
 	\centerline{
	\includegraphics[width=0.377\textwidth]{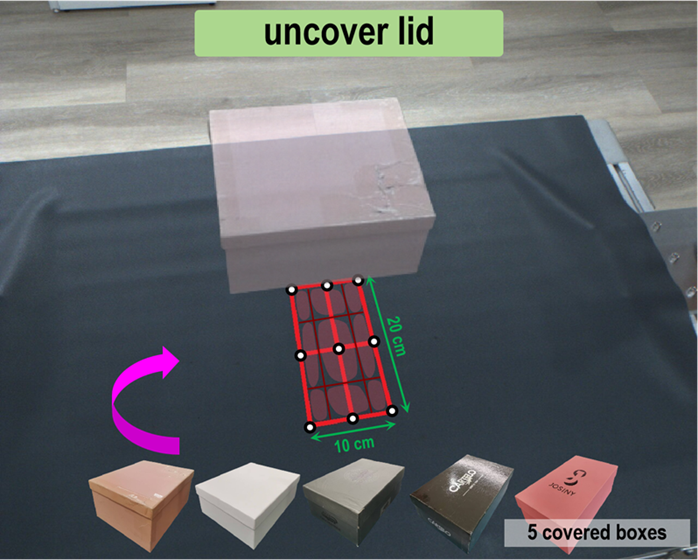}
 	\includegraphics[width=0.377\textwidth]{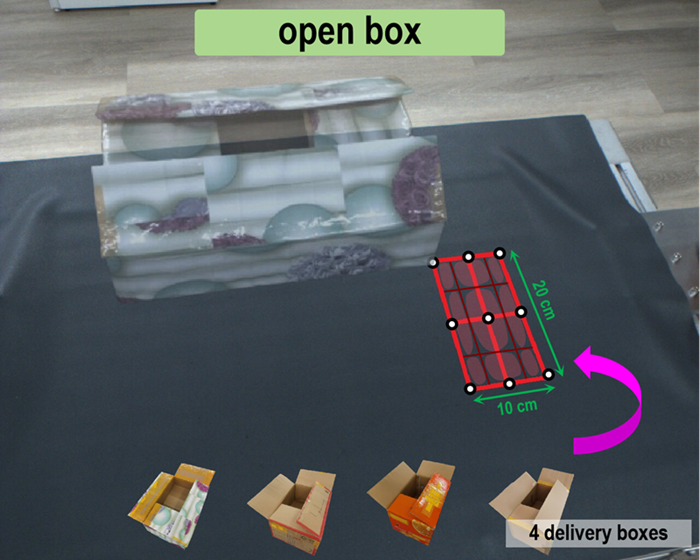}
	}
	\caption{Illustration of the effective placement range of manipulated objects on the workbench for each task during training and testing. We mainly show the predefined position points of the objects in the automatic verification phase (indicating by black-white dots), as well as the approximate distribution of the object positions after using the geometry transformation (indicating by spread translucent red scopes). Best to view after zooming in.}
	\label{suppAreas}
	\vspace{-10pt}
\end{figure*}

\begin{figure*}[]
	\centering
 	\includegraphics[width=\textwidth]{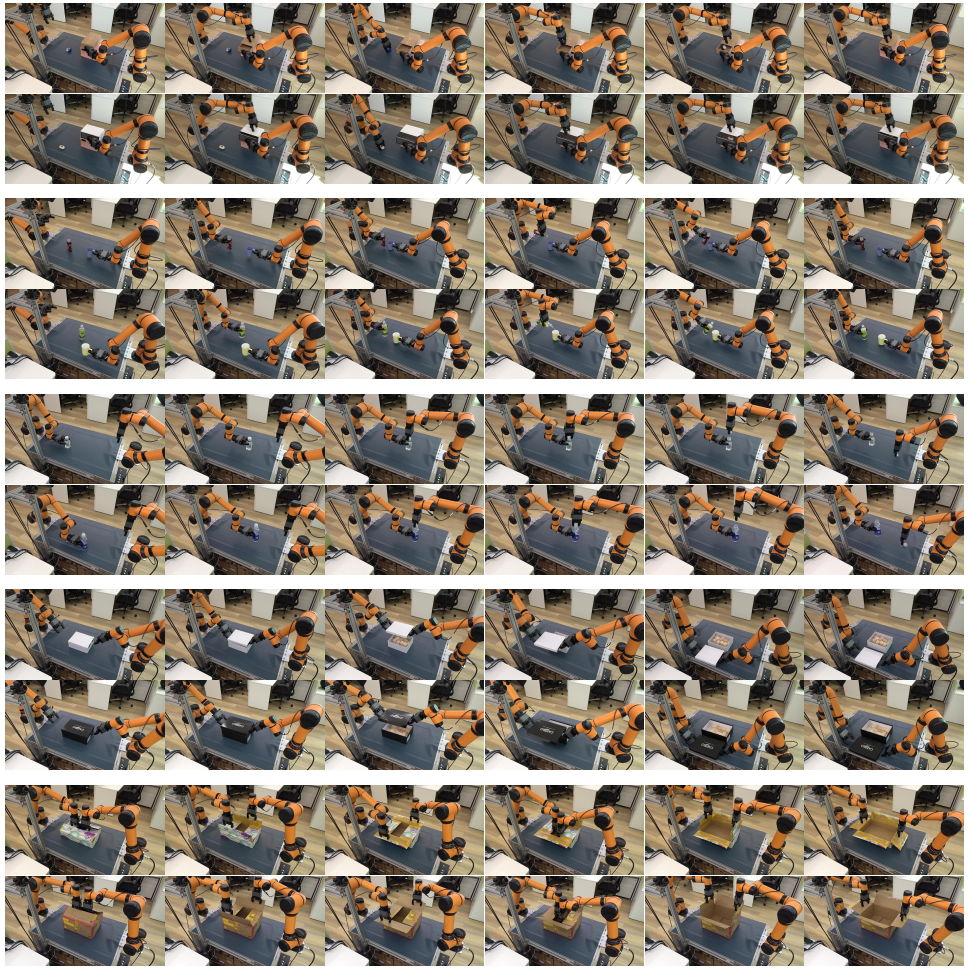}
	\caption{Qualitative rollout samples from the third-person perspective for all real robot evaluation scenarios. From top to bottom, they are the five long-horizon bimanual tasks mentioned in the main paper. Best to view after zooming in.}
	\label{moreRollouts}
	\vspace{-10pt}
\end{figure*}

\begin{figure*}[]
	\centering
	\includegraphics[width=\textwidth]{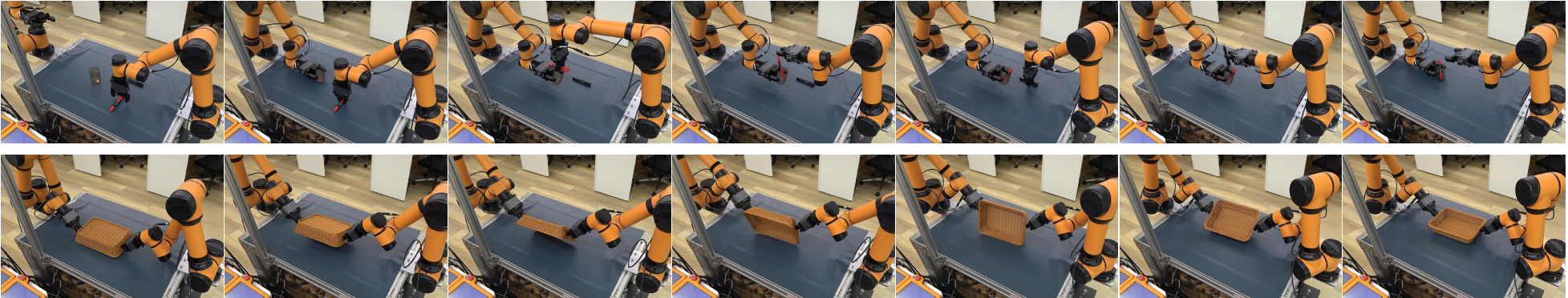}
	\caption{Rollout examples of another two new tasks including \texttt{reorient pen} (top row) and  \texttt{flip basket} (bottom row).}
	\label{addRollouts}
	\vspace{-10pt}
\end{figure*}

\begin{figure*}[]
	\centering
	\includegraphics[width=\textwidth]{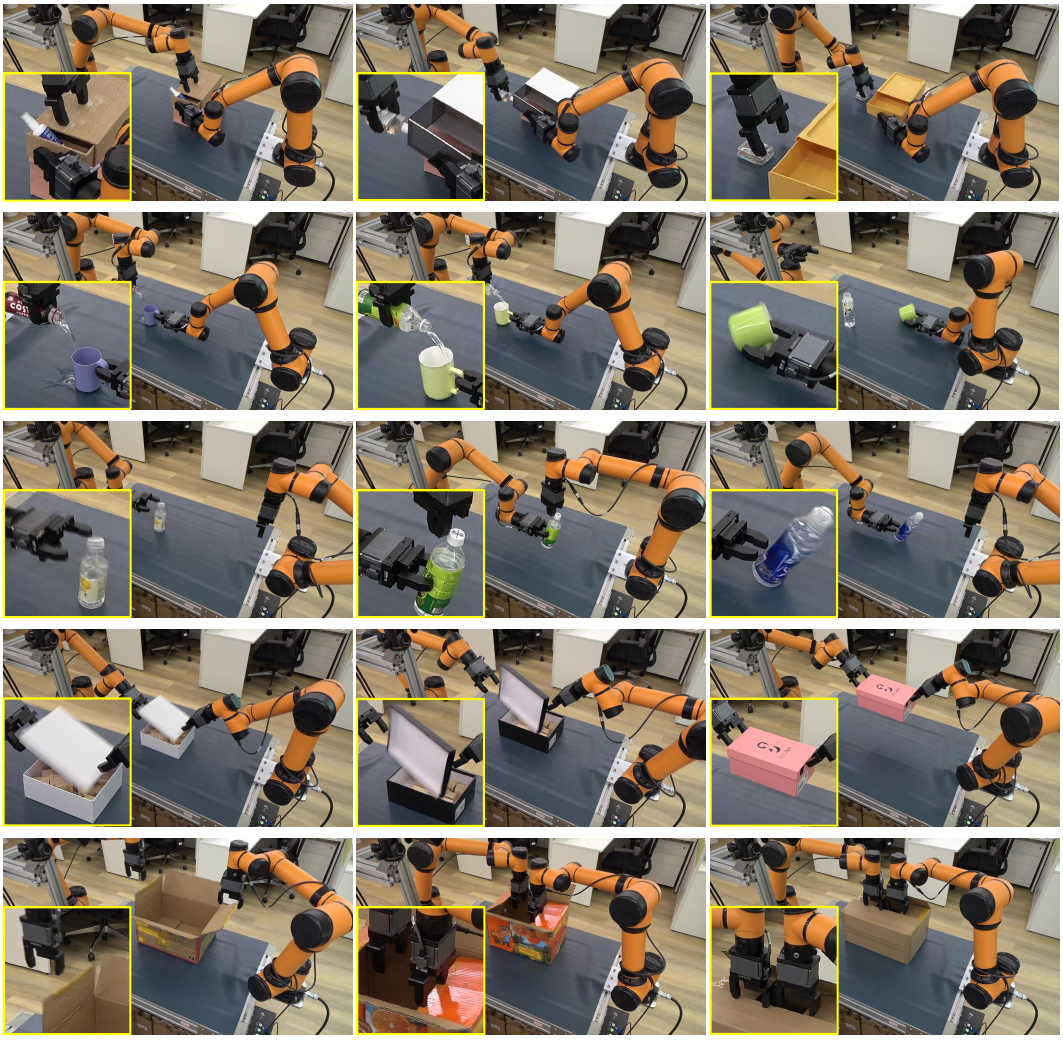}
	\caption{From top to bottom, we have examples of failed cases in all five tasks during evaluation. We have outlined and magnified the areas where the failures occurred so that we can quickly examine them. Best to view after zooming in.}
	\label{failedCases}
	\vspace{-10pt}
\end{figure*}

\subsection{Details of Our Selected Bimanual Tasks}

We here summarize details related to all manipulation tasks, including object size, hand trajectory visualization, number of keyframes and the valid manipulation area.

\subsubsection{Real size of manipulated objects}
In order to give readers a more intuitive cognition of the size of all the objects used in our experiment, we have summarized the centimeter-level shape information of the objects involved in detail in Fig.~\ref{suppAssets}. Without loss of generality, we roughly abstract all objects into two typical geometric shapes, namely \textbf{cuboids} (represented by length, width and height) and \textbf{cylinders} (represented by height and diameter). The cuboids include 9 everyday objects and 3 drawers in the \texttt{pull drawer} task, 5 covered boxes in the \texttt{uncover lid} task, and 4 express boxes in the \texttt{open box} task. The cylinders include the 6 bottles and 3 mugs in the \texttt{pour water} task, and the 6 caps in the \texttt{unscrew bottle} task. In particular, we also counted the height of the lid in task \texttt{uncover lid} and the length of the flippable wings on both sides in task \texttt{open box}. These objects are all within the size range that the gripper can grasp and the operating table can place. We expect that these detailed statistics can help better understand and reproduce each task.

\subsubsection{Hand motion extraction results}
The movement mode of all bimanual tasks we designed mainly comes from a single-shot teaching of both human hands. In Fig.~\ref{tasksTrajs}, we show the detailed visualization results of extracted hand motion trajectories so that we can better understand the entire process of manipulation, including which objects each arm contacts and the order of motion. Our subsequent demonstration proliferation and learned action prediction policies will follow the same motion pattern.

\subsubsection{Determination of keyframes in each task}
As described in the main text, we use discrete keyframes (\textit{a.k.a.} keyposes) to simplify and represent each long-horizon task as in C2FARM \cite{james2022coarse} and PerAct \cite{shridhar2023perceiver}. Keyframes can be auto-extracted using simple heuristics, such as a change in the open/close end-effector state or local extrema of velocity/acceleration. This abstraction way is extremely effective for the first three strictly asynchronous tasks, but the latter two tasks that require the synchronization of both arms do not perform well due to very few remaining keyframes. Therefore, we artificially increase sampling frames between keyframes with larger step spans to make the manipulation action more stable and smooth. This constraint can also be added to heuristic rules to complete automatic keyframes extraction.

\subsubsection{Effective area on the workbench} Since we used two fixed manipulators, the accessible space is limited. Specifically, as shown in Fig.~\ref{suppAreas}, we marked the distribution of the effective areas where the objects were located in each task on the table platform. In the training demonstrations we collected and the real robot evaluation phase, we would not place objects outside these areas to avoid exceeding the reach limit of two arms. Note that this does not mean that the policies we trained do not generalize to different locations. Even in a restricted area, the position and orientation of the manipulated objects during testing may be completely new, so it is still a non-trivial out-of-distribution (OOD) situation.

In Fig.~\ref{suppAreas}, we use a series of black-white dots to represent each pre-fixed point, which corresponds to the position of the manipulated object in the auto-rollout phase. These points are defined relative to the area where the object touches the table, and can be at its geometric center (\texttt{open box}), the midpoint of a side (\texttt{uncover lid}), or a corner in a fixed direction (\texttt{open box}, \texttt{pour water}, \texttt{unscrew bottle} and \texttt{open box}). We show some examples of semi-transparent manipulated objects in each sub-image. The effective range drawn on the table also refers to these proxy points. Therefore, considering that the object itself may be quite large (refer Fig.~\ref{suppAssets}), the area it actually covers will be much larger. On the other hand, we utilize the semi-transparent red area to indicate the approximate distribution of objects after their positions have been modified using the point cloud-based geometric transformation. We apply measured parameters to control the augmented objects to basically cover the entire valid area without overlapping, so that the trained model has robust generalization of position variations. Moreover, this design ensures that the absolute displacement of the object will not be too large, thus avoiding obvious violations of the imaging rules under perspective projection.

\subsection{Evaluation Results and Performance Analysis}

\subsubsection{More qualitative examples}
In Fig.~\ref{moreRollouts}, we show qualitative rollout samples from the third-person perspective for all evaluation tasks we mentioned in the main paper. We still choose images near the keyframes to illustrate more effectively, e.g., the model’s generalization to object \textit{category} and \textit{location} variations. These examples show more complete scenes and the motion of two robot arms, and can be considered as a supplement to the limited field of view of the binocular observation camera. Note that these third-person video recordings do not participate in any training and testing.

Specifically, from top to bottom in Fig.~\ref{moreRollouts}, we have: (1.1) open the \textit{gray wooden drawer}, pick up the \textit{vaseline box} and put it into the drawer, and close the drawer; (1.2) open the \textit{white plastic drawer}, pick up the \textit{headphone case} and put it into the drawer, and close the drawer; (2.1)\&(2.2) grasp and pick up the \textit{mug cup}, grasp and pick up the \textit{drink bottle}, pour the water contained in the bottle into the mug cup, and place back the mug cup and bottle; (3.1)\&(3.2) grasp and pick up the \textit{drink bottle}, unscrew the \textit{circular cap} of the bottle, and place back the bottle and cap; (4.1)\&(4.2) use two arms to close to the lid of the \textit{covered box}, lift up the box lid, and place down the box lid; (5.1)\&(5.2) use two arms to close to the two upper longer wings of the \textit{delivery box}, open the wings, close to the another two lower shorter wings, and open the wings. For more intuitive qualitative results, please refer to the recorded videos.

\subsubsection{Rollouts of two new tasks}
We show in the main paper two additional tasks that can be injected into dual-arm manipulators after a single-shot teaching following the proposed YOTO paradigm. Fig.~\ref{addRollouts} shows a series of third-person recordings of real rollouts. They involve two important atomic skills: \textit{reorientation} and \textit{rearrangement}. The two tasks are: (1) \texttt{reorient pen}. Pick up two pens placed in different directions and place them in a cup. (2) \texttt{flip basket}. Flip the non-prehensile woven basket with the mouth facing down 180 degrees so that it faces upwards.

\subsubsection{Analysis of failure cases}
Although our method BiDP outperforms many strong baselines \cite{zhao2023learning, chi2023diffusion, ze2024dp3, yange2024quibot} for addressing long-horizon bimanual manipulation tasks, it still presents various failure cases during evaluation. Below, we focus our analysis on execution failure of our BiDP in real-world experiments. In Fig.~\ref{failedCases}, we show some representative failure examples of all real robot executions we have performed with our method.

Specifically, from top to bottom, we have collected failed cases like: (1.1) The object was \underline{wrongly placed} in the drawer so that the drawer could not be closed successfully; (1.2) The object \underline{collided} with the drawer during the transfer process, causing the drawer to move out of position and affecting its closing operation; (1.3) A \underline{grasping error} occurred during the object picking process, causing the object to fail to be picked up; (2.1) The mouth of the bottle is \underline{not aligned} with the mouth of the mug cup, causing more water to spill out; (2.2) The \underline{contact point was biased} when grasping the mug handler, making it easy for water to spill out; (2.3) A \underline{grabbing error} occurred while picking up the mug causing it to fall over; (3.1) When picking up the bottle, the gripper \underline{squeezed the bottle cap}, causing the grip to fail; (3.2) The gripper \underline{fails to clamp the center} of the bottle cap, causing the cap to fail to be twisted off; (3.3) When picking up the bottle, the gripper \underline{collided} with the bottle body, causing the bottle to fall and the gripping failed; (4.1) The box lid fell off due to \underline{lack of coordination} when opening it; (4.2) The box lid fell off due to \underline{lack of coordination} when transferring it. (4.3) \underline{Inappropriate distance between arms} caused that the entire covered box was lifted and moved without opening the box lid; (5.1) After the delivery box was fully opened, due to \underline{a defect in the pressing action} (such as not long enough or deep enough), a wing rebounded back due to the tension at the hinges; (5.2) Due to the delivery box \underline{displacement or inaccurate action prediction}, a lower shorter wing could not be successfully opened. (5.3) Due to \underline{inaccurate action prediction}, a upper longer wing failed to open successfully. These failure cases point out the direction that needs further exploration in the future. For more intuitive qualitative results, please refer to our recorded videos.


\end{document}